\documentclass[a4paper,12pt]{article}

\usepackage{comment}
\usepackage{fancyvrb}
\usepackage{graphicx}
\usepackage[ragged]{footmisc}
\usepackage{ragged2e}
\usepackage{tipa} % International Phonetics Alphabet
\usepackage{wasysym} % LaTeX support file to use the WASY2 fonts
\usepackage{gensymb} % Generic symbols for both text and math mode
\usepackage{geometry}
\usepackage{hyperref}

\begin{document}

\title{Proposal for the creation of a research facility for the development of the SP machine\protect\footnote{An electronic copy of this document may be downloaded via \href{http://bit.ly/1zZjjIs}{bit.ly/1zZjjIs}.}}

\author{J Gerard Wolff\footnote{Dr Gerry Wolff, BA (Cantab), PhD (Wales), CEng, MIEEE, MBCS (CITP); CognitionResearch.org, Menai Bridge, UK; \href{mailto:jgw@cognitionresearch.org}{jgw@cognitionresearch.org}; +44 (0) 1248 712962; +44 (0) 7746 290775; {\em Skype}: gerry.wolff; {\em Web}: \href{http://www.cognitionresearch.org}{www.cognitionresearch.org}.} \and Vasile Palade\footnote{Dr Vasile Palade, MEng (Tech.~Univ.~of Bucharest),  PhD (University of Gala\c{t}i), IEEE Senior Member; Reader in Pervasive Computing, Department of Computing, Coventry University, UK; \href{mailto:vasile.palade@coventry.ac.uk}{vasile.palade@coventry.ac.uk}; +44 (0) 2477 659190; +44 (0) 7912 043982; {\em Skype}: v\_palade; {\em Web}: \href{http://bit.ly/1xNM5xr}{bit.ly/1xNM5xr}, \href{http://bit.ly/1Buyk9E}{bit.ly/1Buyk9E}.}}

\maketitle

\begin{abstract}

This is a proposal to create a research facility for the development of a high-parallel version of the {\em SP machine}, based on the {\em SP theory of intelligence}. We envisage that the new version of the SP machine will be an open-source software virtual machine, derived from the existing {\em SP computer model}, and hosted on an existing high-performance computer. It is intended as a means for researchers everywhere to explore what can be done with the system and to create new versions of it. The SP system is a unique attempt to simplify and integrate observations and concepts across artificial intelligence, mainstream computing, mathematics, and human perception and cognition, with information compression via the matching and unification of patterns as a unifying theme. Many potential benefits and applications flow from this simplification and integration. These include potential with problems associated with big data; potential to facilitate the development of autonomous robots; and potential in other areas including: unsupervised learning, natural language processing, several kinds of reasoning, fuzzy pattern recognition and recognition at multiple levels of abstraction, computer vision, best-match and semantic forms of information retrieval, software engineering, medical diagnosis, simplification of computing systems, and the seamless integration of diverse kinds of knowledge and diverse aspects of intelligence. Additional motivations for creating the proposed facility include the potential of the SP system to help solve problems in defence, security, and the detection and prevention of crime; the potential impact of the development in terms of economic, social, environmental, and academic criteria, and in terms of publicity; and the potential for international influence in research. The main elements of the proposed facility are described, including support for the development of {\em SP-neural}, a neural version of the SP machine. The facility should be permanent in the sense that it should be available for the foreseeable future, and it should be designed to facilitate its use by researchers anywhere in the world.

\end{abstract}

\section{Introduction}\label{introduction_section}

\begin{quote}

``There is nothing more practical than a good theory'', Kurt Lewin \cite[p.~169]{lewin_1952}.

\end{quote}

% Dan yr Ogof analogy;

This is a proposal to create a high-parallel version of the {\em SP machine}, which we will refer to as ``SPM'', and a research facility for its further development, which we will refer to as ``SPF''. It is envisaged that SPM will be created as an open-source software virtual machine, derived from the existing {\em SP computer model} (SP71), and hosted on an existing high-performance computer. SPF is intended to be a means for researchers everywhere to explore what can be done with the SP system and to create new versions of it.

\subsection{The SP theory and the SP machine}\label{theory_and_machine_section}

The basis of the proposed development is the {\em SP theory of intelligence}. This is a unique attempt to simplify and integrate observations and concepts across artificial intelligence, mainstream computing, mathematics, and human perception and cognition, with information compression via the matching and unification of patterns as a unifying theme.\footnote{The name ``SP'' is short for {\em Simplicity} and {\em Power}, because compression of any given body of information, {\bf I}, may be seen as a process of reducing ``redundancy'' of information in {\bf I} and thus increasing its ``simplicity'', whilst retaining as much as possible of its non-redundant descriptive and explanatory ``power''.}

Many potential benefits and applications flow from this simplification and integration, as detailed in Section \ref{motivations_section} and Appendix \ref{potential_benefits_applications_appendix}. Several of those benefits and applications may be realised on relatively short timescales, as indicated in the introduction to Appendix \ref{potential_benefits_applications_appendix}.

The SP theory was not dreamed up overnight. It is the product of about 20 years of development and testing.

A central idea in the SP theory is the powerful concept of {\em multiple alignment}, borrowed and adapted from that concept in bioinformatics, and outlined in Appendix \ref{ma_appendix}. It appears that multiple alignment provides a key to versatility and adaptability in intelligent systems.

Expressing the theory in a computer model has helped to reduce vagueness in the theory, it has provided a means of testing candidate ideas---many of which have been rejected as a result of this testing---and it is a means of demonstrating what can be done with the system.

There is an outline of the SP theory and the SP computer model in Appendix \ref{outline_of_sp_theory_and_model_appendix}, with pointers to where fuller information may be found. Distinctive features and apparent advantages of the SP system are summarised in Appendix \ref{distinctive_features_and_advantages_appendix}. Some of the potential benefits and applications of the SP system are summarised in Section \ref{motivations_section} and Appendix \ref{potential_benefits_applications_appendix}, with pointers to more comprehensive sources of information.

The wide scope of the SP system, and the wide range of its potential benefits and applications, means that there are many more avenues to be explored than could be tackled by any one research group. Hence the need for SPF. It is intended to facilitate large-scale collaboration, very much in the spirit of:

\begin{quote}

``The creation of this new era of [cognitive] computing is a monumental endeavor ... no company can take on this challenge alone. So we look to our clients, university researchers, government policy makers, industry partners, and entrepreneurs---indeed the entire tech industry---to take this journey with us.'' \cite[Preface]{kelly_hamm_2013}.

\end{quote}

\noindent and

\begin{quote}

``... fund activities that support integration and collaboration within the research community; for instance, CPPs [Collaborative Computational Projects], consortia, networks, international interactions.''\footnote{From p.~9 in {\em EPSRC E-Infrastructure Roadmap}, August 2014, \href{http://bit.ly/1d8Eh3I}{bit.ly/1d8Eh3I}.}

\end{quote}

\subsection{Presentation}

In what follows, we'll first describe some motivations for creating SPF (Section \ref{motivations_section}). Then we'll describe the facility itself and how it may be used (Section \ref{hpsp_section}). There is an estimate of costs in Section \ref{costs_section}, and concluding remarks in Section \ref{conclusion_section}.

\section{Motivations}\label{motivations_section}

In terms of policies of the UK Government and the UK Research Councils, SPM and SPF tick several boxes, and there are other reasons for creating SPF and developing SPM. These motivations are described in subsections below.

\subsection{The power and potential of multiple alignment}\label{broad_scope_section}

Probably the most distinctive feature of the SP system is the powerful concept of multiple alignment, as it has been adapted from bioinformatics (Appendix \ref{distinctive_features_and_advantages_appendix}). In broad terms, this concept provides a key to:

\begin{itemize}

\item {\em Simplification of concepts and systems}. Diverse concepts in computing and cognition may be assimilated to the relatively simple concept of multiple alignment (\cite{wolff_2006}, \cite[Section2 and 4]{sp_benefits_apps}), and that concept can mean substantial simplifications of computing systems (\cite[Section 4.4.2]{wolff_2006}, \cite[Section 5]{sp_benefits_apps}).

\item {\em Versatility in intelligence}. The SP system has strengths in unsupervised learning, natural language processing, pattern recognition, information retrieval, several kinds of reasoning, planning, problem solving, information compression, and more \cite{wolff_2006,sp_extended_overview}.

\item {\em Adaptability in intelligence}. The SP system's strengths in unsupervised learning \cite[Chapter 9]{wolff_2006} may provide a foundation for other kinds of learning \cite[Section V]{sp_autonomous_robots}. Via unsupervised learning, and perhaps via other kinds of learning, the system has potential to promote human-like adaptability in intelligence.

\item {\em Seamless integration of structures and functions}. The SP system, with multiple alignment centre stage, has potential to promote seamless integration of diverse kinds of knowledge and diverse aspects of intelligence \cite[Section 7]{sp_benefits_apps}, an integration that appears to be essential for effective reasoning and problem solving in intelligent systems.

\end{itemize}

\noindent More specifically, there are many potential benefits and applications that flow from the multiple alignment concept, described in outline in Appendix \ref{potential_benefits_applications_appendix} and in more detail in \cite{wolff_2006,sp_benefits_apps,sp_big_data,sp_autonomous_robots,wolff_sp_intelligent_database,wolff_medical_diagnosis}. As noted in Section \ref{benefits_applications_section}, the potential value, worldwide, of those benefits and application, has been estimated to be at least \$190 billion, {\em every year}.

The estimated cost of setting up SPF is $\pounds 634,700$ and recurrent costs are estimated to be $\pounds 173,800$ pa (Section \ref{costs_section}).

Taking account of these estimated costs, and the potential benefits and applications, it appears that the proposed facility would be a good investment.

\subsection{UK industrial strategy}\label{uk_industrial_strategy_section}

The development of SPF would chime well with the goal of ``supporting emerging technologies'', one of five main themes in the UK's industrial strategy.\footnote{See {\em Industrial strategy: government and industry in partnership}, {\em GOV.UK}, 2013-08-06, \href{http://bit.ly/1kYdtBS}{bit.ly/1kYdtBS}.}

More specifically, the SPF fits well with the Government's ``information economy strategy'',\footnote{See {\em Information Economy Strategy} (PDF), June 2013, \href{http://bit.ly/1d5tn8h}{bit.ly/1d5tn8h}.} and its focus on ``eight `great technologies' which will propel the UK to future growth'':\footnote{See {\em $\pounds 600$ million investment in the eight great technologies}, Department for Business, Innovation \& Skills, 2013-01-24, \href{http://bit.ly/QFwhwi}{bit.ly/QFwhwi}; and ``The `eight great technologies' which will propel the UK to future growth receive a funding boost'', speech given by the Rt Hon David Willetts MP, former Minister for Universities and Science, {\em GOV.UK}, 2013-01-23, \href{http://bit.ly/1wKCqII}{bit.ly/1wKCqII}.}

The SP system has considerable potential to contribute to development of two of those ``great technologies''---{\em big data} (discussed in the next subsection) and {\em robotics and autonomous systems} (discussed in Section \ref{robotics_autonomous_section}). The SP system is also relevant to development in a third area---{\em energy}---discussed briefly in Section \ref{energy_section}).

In addition, the very broad scope of the SP system (Section \ref{broad_scope_section}), and the provision of a facility for researchers anywhere in the world (Section \ref{open_to_researchers_everywhere_section}), is likely, via synergies of various kinds, encourage ``better science'', ``new kinds of science'', and ``interdisciplinary projects'', in keeping with the Government's {\em A Strategic Vision for UK e-Infrastructure}.\footnote{Report from the Department for Business, Innovation and Skills, January 2012, \href{http://bit.ly/1FrAQzW}{bit.ly/1FrAQzW}, p.~21.}

\subsection{Big data}\label{big_data_section}

Outlined here are some significant challenges associated with big data and how the SP system may help to overcome them.

\subsubsection{Challenges}\label{challenges_section}

There is a tendency amongst some people to write or speak as if big data was an unalloyed blessing and that all the necessary technologies are available to exploit it. But it is clear from the book {\em Smart Machines} by John Kelly and Steve Hamm (both of IBM) \cite{kelly_hamm_2013}, and also from {\em Frontiers in Massive Data Analysis} from the US National Research Council \cite{national_research_council_2013}, that much of the potential value of big data cannot be realised currently because of significant unsolved problems in that area.

It is also clear from {\em Smart Machines} that significant problems associated with big data cannot be solved by tinkering with existing designs for computers. Radical rethinking of the architecture of computers will be needed, probably drawing on what we can learn about the workings of the human brain.

And achieving this transition to ``cognitive computing'' will require large-scale collaboration, as indicated in Section \ref{theory_and_machine_section}.

\subsubsection{Potential solutions}\label{potential_solutions_section}

SPM and SPF are relevant to the issues just described because: the SP system has been developed in the spirit of cognitive computing, drawing heavily on research on human perception and cognition, and neuroscience; it offers radical solutions to problems posed by big data; and the existence of SPF would encourage and facilitate large-scale collaboration amongst researchers all around the world.

The paper {\em Big data and the SP theory of intelligence} \cite{sp_big_data} describes how the SP system may help to solve nine of the problems associated with big data, as outlined in Appendix \ref{big_data_appendix}. Three of those problems, and their potential solutions, are highlighted here:

\begin{itemize}

\item {\em The problem of variety in big data}. There is a pressing need in computing to tame the great variety of formalisms and formats for the representation of knowledge, each with their own mode of processing. The multiple alignment framework in the SP system has clear potential as a {\em universal framework for the representation and processing of diverse kinds of knowledge} (UFK) \cite[Section III]{sp_big_data}.

\item {\em Energy consumption}. The SP system, in conjunction with the concept of ``data-centric computing'' \cite[Chapter 5]{kelly_hamm_2013}, has potential to ``make computers many orders of magnitude more energy efficient'' \cite[p.~88]{kelly_hamm_2013} via two key principles \cite[Section IX]{sp_big_data}: 1) Taking advantage of statistical information that the system gathers as a by-product of how it works; 2) Cutting out much searching by making direct connections between ``neural symbols'' in ``SP-neural'' (Appendix \ref{sp-neural_appendix}).

\item {\em Transmission of information}. The SP system has potential to yield dramatic economies in the transmission of information, partly by making big data smaller \cite[Section VII]{sp_big_data}, but perhaps more importantly via analysis/synthesis \cite[Chapter 18]{sayood_2012}, based on the judicious separation of {\em encoding} and {\em grammar} \cite[Section VIII]{sp_big_data}.

\end{itemize}

As noted in Appendix \ref{big_data_appendix}, it appears that, considering the nine proposed solutions collectively, and in several cases individually, there are no alternatives that can rival the potential of what is described in \cite{sp_big_data}.

With regard to the Government's information economy strategy,\footnote{{\em Information Economy Strategy} (PDF), June 2013, \href{http://bit.ly/1d5tn8h}{bit.ly/1d5tn8h}.} ``The UK now has the opportunity to take a lead in the global efforts to deal with the volume, velocity and variety of data created each day. This will require continued infrastructure investment ...'' (p.~7). SPF has potential to be a major plank in that infrastructure, and at modest cost (Section \ref{costs_section}).

\subsection{Robotics and autonomous systems}\label{robotics_autonomous_section}

As mentioned above, {\em robotics and autonomous systems} is another of the Government's ``eight great technologies'' where the SP system may make a contribution---perhaps via the UK Robotics and Autonomous Systems Network (UK-RAS Network)\footnote{See, for example, ``Widespread backing for UK robotics network'', The Engineer, 2015-06-24, \href{http://bit.ly/1BR9qE2}{bit.ly/1BR9qE2}.} or otherwise---and the potential is considerable.

In the paper {\em Autonomous robots and the SP theory of intelligence} \cite{sp_autonomous_robots}, robots are ``autonomous'' if they do not depend on external intelligence or power supplies, and are mobile. It is also assumed that they are designed to exhibit as much human-like intelligence as possible. The paper describes how the SP system may contribute in three main areas:

\begin{itemize}

\item {\em Computational efficiency, the use of energy, and the size and weight of computers}. ``To field a conventional computer with [human-like] cognitive capacity would require gigawatts of electricity and a machine the size of a football field.'' \cite[p.~75]{kelly_hamm_2013}. If a robot is to be autonomous as described above, it needs a `brain' that is efficient enough to do all the necessary processing without external assistance, does not require an industrial-scale power station to meet its energy demands, and is small enough and light enough to be carried around---things that are difficult or impossible to achieve with current technologies.

    Section III of \cite{sp_autonomous_robots} describes how these things may be achieved in a revised and updated version of arguments in \cite[Section IX]{sp_big_data}. The SP system may help: by reducing the size of data to be processed; by exploiting statistical information that the system gathers as an integral part of how it works; and via a revised version of Donald Hebb's \cite{hebb_1949} concept of a ``cell assembly''.

\item {\em Towards human-like versatility in intelligence}. If a robot is to operate successfully in an environment where people cannot help, or where such opportunities are limited, it needs as much as possible of the versatility in intelligence that people may otherwise provide.

    The SP system demonstrates versatility via its strengths in areas such as unsupervised learning, natural language processing, fuzzy pattern recognition and recognition at multiple levels of abstraction, best-match and semantic forms of information retrieval, several kinds of reasoning, planning, problem solving, information compression, and more \cite[Section IV]{sp_autonomous_robots}.

    But the SP system is not simply a kludge of different AI functions. Owing to its focus on simplification and integration of concepts in computing and cognition (Appendix \ref{outline_of_sp_theory_and_model_appendix}), it promises to reduce or eliminate unnecessary complexity and to avoid awkward incompatibilities between poorly-integrated subsystems, as indicated in Appendix \ref{simplification_integration_concepts_appendix}.

\item \sloppy {\em Towards human-like adaptability in intelligence}. The SP system's strengths in unsupervised learning and other aspects of intelligence may help to achieve human-like adaptability in intelligence via: one-trial learning; the learning of natural language; learning to see; building 3D models of objects and of a robot's surroundings; learning regularities in the workings of a robot and in the robot's environment; exploration and play; learning major skills; and learning via demonstration \cite[Section V]{sp_autonomous_robots}.

\end{itemize}

This new approach to the development of autonomous robots has several distinctive features (Appendix \ref{distinctive_features_and_advantages_appendix}), especially the concept of multiple alignment. It appears that this approach has considerable potential to promote computational and energy efficiency in a robot's brain, and to achieve human-like versatility and adaptability in intelligence.

\subsection{Energy}\label{energy_section}

The SP system may also contribute indirectly to another of the areas identified as one of the ``eight great technologies''---{\em energy}---via increases in efficiency in the workings of computers and in the transmission of data, as outlined in Sections \ref{big_data_section} and \ref{robotics_autonomous_section}, and described more fully in \cite[Sections VIII and IX]{sp_big_data} and also in \cite[Section III]{sp_autonomous_robots}.

\subsection{Other potential benefits and applications of the SP system}\label{benefits_applications_section}

In addition to big data (Section \ref{big_data_section}), robotics and autonomous systems (Section \ref{robotics_autonomous_section}), and energy (Section \ref{energy_section}), there are several other potential benefits and applications of the SP system, outlined in Appendix \ref{potential_benefits_applications_appendix}. As noted there, several of these potential benefits and applications may be realised on relatively short timescales with existing high-performance computers or even ordinary computers.

Peer-reviewed papers about potential benefits and applications that have not already been mentioned include \cite{sp_vision} (application of the SP theory to the understanding of natural vision and the development of computer vision), \cite{wolff_medical_diagnosis} (application of the SP system to medical diagnosis), and \cite{wolff_sp_intelligent_database} (the SP system as an intelligent database), and \cite{sp_benefits_apps} (several other potential benefits and applications including unsupervised learning, natural language processing, pattern recognition, software engineering, information compression, the semantic web, bioinfomatics, data fusion, simplification of computing systems, and the seamless integration of structures and functions in diverse kinds of knowledge).

In view of the wide scope of the SP system, and evidence of its potential, it seems reasonable to estimate that it could add at least 5\% to the value of IT investments, worldwide. Since these are about \$3.8 trillion annually,\footnote{See, for example, ``Gartner: Big data will help drive IT spending to \$3.8 trillion in 2014'', {\em InfoWorld}, 2013-01-03, \href{http://bit.ly/Z00SBr}{bit.ly/Z00SBr}.} the value of the SP concepts, {\em every year}, would be at least \$190 billion! \cite[Section 8]{sp_benefits_apps}.

\subsection{Defence, security, and the detection and prevention of crime}\label{security_section}

With regard to ``Security in a changing world'', part of the ``vision'' \cite[p.~13]{stfc_2010} of the UK's Science \& Technology Facilities Council (STFC), several of the strengths of the SP system that have been mentioned are relevant to defence, security, and the detection and prevention of crime. These include:

\begin{itemize}

\item Big data (Section \ref{big_data_section}), and data fusion \cite[Section 6.10.5]{sp_benefits_apps}.

\item Robotics and autonomous systems (Section \ref{robotics_autonomous_section}).

\item Natural language processing \cite[Chapter 5]{wolff_2006}, \cite[Section 6.2]{sp_benefits_apps}.

\item Pattern recognition (see Appendix \ref{pattern_recognition_appendix} and \cite[Chapter 6]{wolff_2006}).

\item The SP system may function as an intelligent database \cite{wolff_sp_intelligent_database}.

\item Computer vision \cite{sp_vision}.

\item Several kinds of reasoning, including one-step `deductive' reasoning, abductive reasoning, reasoning with probabilistic networks and trees, reasoning with if-then rules, nonmonotonic reasoning, `explaining away', and causal diagnosis  \cite[Chapter 7]{wolff_2006}.

\item Planning and problem solving \cite[Chapter 8]{wolff_2006}.

\item Unsupervised learning \cite[Chapter 9]{wolff_2006}, \cite[Section 6.1]{sp_benefits_apps}.

\item The potential of the system to simplify computing system \cite[Sections 4 and 5]{sp_benefits_apps}.

\item The potential of the system to promote seamless integration of diverse forms of knowledge and diverse aspects of intelligence \cite[Section 7]{sp_benefits_apps}.

\end{itemize}

This last point is particularly relevant to forensic work and security work, because of the need to remove artificial barriers between different aspects of intelligence, different kinds of knowledge, and different kinds of processing, so that all kinds of intelligence and knowledge can be brought to bear in solving crimes and in the prevention of acts of terrorism. Similar things may be said about autonomous robots that aspire to human-like versatility and adaptability in intelligence.

\subsection{Contributing to the work of the planned Alan Turing Institute for Data Science and the planned Cognitive Computing Research Centre}\label{alan_turing_institute_section}

\begin{quote}

``The Alan Turing Institute for Data Science will benefit from a $\pounds42$ million government investment over 5 years that will strengthen the UK's aims to be a world leader in the analysis and application of big data. It will also ensure that the UK is at the forefront of data-science in a rapidly moving, globally competitive area, enabling first-class research in an environment that brings together theory and practical application.''\footnote{See ``Plans for world class research centre in the UK'', {\em GOV.UK}, 2014-03-18, \href{http://bit.ly/1j6CZYB}{bit.ly/1j6CZYB}.}

\end{quote}

The SP system and SPF have potential to contribute to the work of the planned {\em Alan Turing Institute for Data Science}, and also the planned {\em Cognitive Computing Research Centre},\footnote{See ``Autumn Statement 2014'', {\em H M Treasury}, December 2014, p.~50, PDF, \href{http://bit.ly/15OO6QS}{bit.ly/15OO6QS}.} by facilitating both the analysis and application of big data (Section \ref{big_data_section}).

\subsection{Impact}\label{impact_section}

\begin{quote}

``The Higher Education Funding Council for England (HEFCE), Research Councils UK (RCUK) and Universities UK (UUK) have a shared commitment to support and promote a dynamic and internationally competitive research and innovation base that makes an increased and sustainable contribution, both nationally and globally, to economic growth, wellbeing, and the expansion and dissemination of knowledge.'', ``Impact Policies'', {\em Research Councils UK}, \href{http://bit.ly/1znJpts}{bit.ly/1znJpts}.

\end{quote}

On five main fronts, the potential impact of SPF is large:

\begin{itemize}

\item {\em Economic}. As noted in Section \ref{benefits_applications_section}, a conservative estimate of the potential value of the SP concepts is at least \$190 billion, every year.

\item {\em Social}. As already noted, there are many potential benefits and applications of the SP system, some of them outlined in Sections \ref{big_data_section} to \ref{benefits_applications_section}, and in Appendix \ref{potential_benefits_applications_appendix}.

\item {\em Environmental}. The SP concepts have potential to deliver more for less---with consequent environmental benefits---via a combination of: 1) gains in economic and social impact (as just noted); 2) reductions in energy consumption and in the size and weight of computers (Sections \ref{potential_solutions_section} and \ref{robotics_autonomous_section}).\footnote{There would also be a need for measures to counteract the effects of ``Jevon's paradox'' ({\em Wikipedia}, \href{http://bit.ly/1dUJF9C}{bit.ly/1dUJF9C}, retrieved 2014-12-22)---how gains in efficiency may, without appropriate controls, lead to an overall increase in consumption.}

\item {\em Academic}. We expect the establishment of SPF, with associated publicity, to encourage and facilitate contributions by research groups and individual researchers all around the world. Given the wide scope of the SP system and the wide range of its potential benefits and applications, the academic impact of SPF is potentially very large. Research that is done with SPF may be seen as a `dividend' from the investment. From the perspective of the UK, research that is done by research groups and individual researchers outside the UK may be seen as a bonus, additional to what would be achieved if the facility were restricted to UK researchers. In addition, there are likely to be benefits arising from synergies and large-scale collaboration (Section \ref{uk_industrial_strategy_section}).

    Initially, we aim to raise awareness of the facility amongst researchers and to encourage relevant research projects. But later, we expect that research with the facility will develop its own momentum.

\item {\em Publicity}. As a condition for using SPF, researchers would be asked to ensure that every relevant publication in an academic journal, collection of academic papers, or conference proceedings, contains an acknowledgement of the facility and of the body or bodies that have provided the necessary funds. This in itself would be publicity. But, in addition, there is potential for publicity in articles in newspapers and magazines, in programmes on radio or TV, and on the internet, in blogs, mailing lists, Facebook pages, and the like.

\end{itemize}

\subsection{International influence}\label{international_influence_section}

\begin{quote}

``Our priorities for maximising international impact are: to seek increased influence for the UK in research, especially in Europe; in the long term to work towards the siting of a major international research facility in the UK.'', \cite[p.~21]{stfc_2010}.

\end{quote}

Given the wide scope of the SP system, the international dimension of SPF, and its potential impact (Section \ref{impact_section}), it has potential to be ``a major international research facility'' and to increase the international influence of STFC or other hosting organisation.

\section{The proposed facility}\label{hpsp_section}

\begin{figure}[!htbp]
\centering
\includegraphics[width=0.9\textwidth]{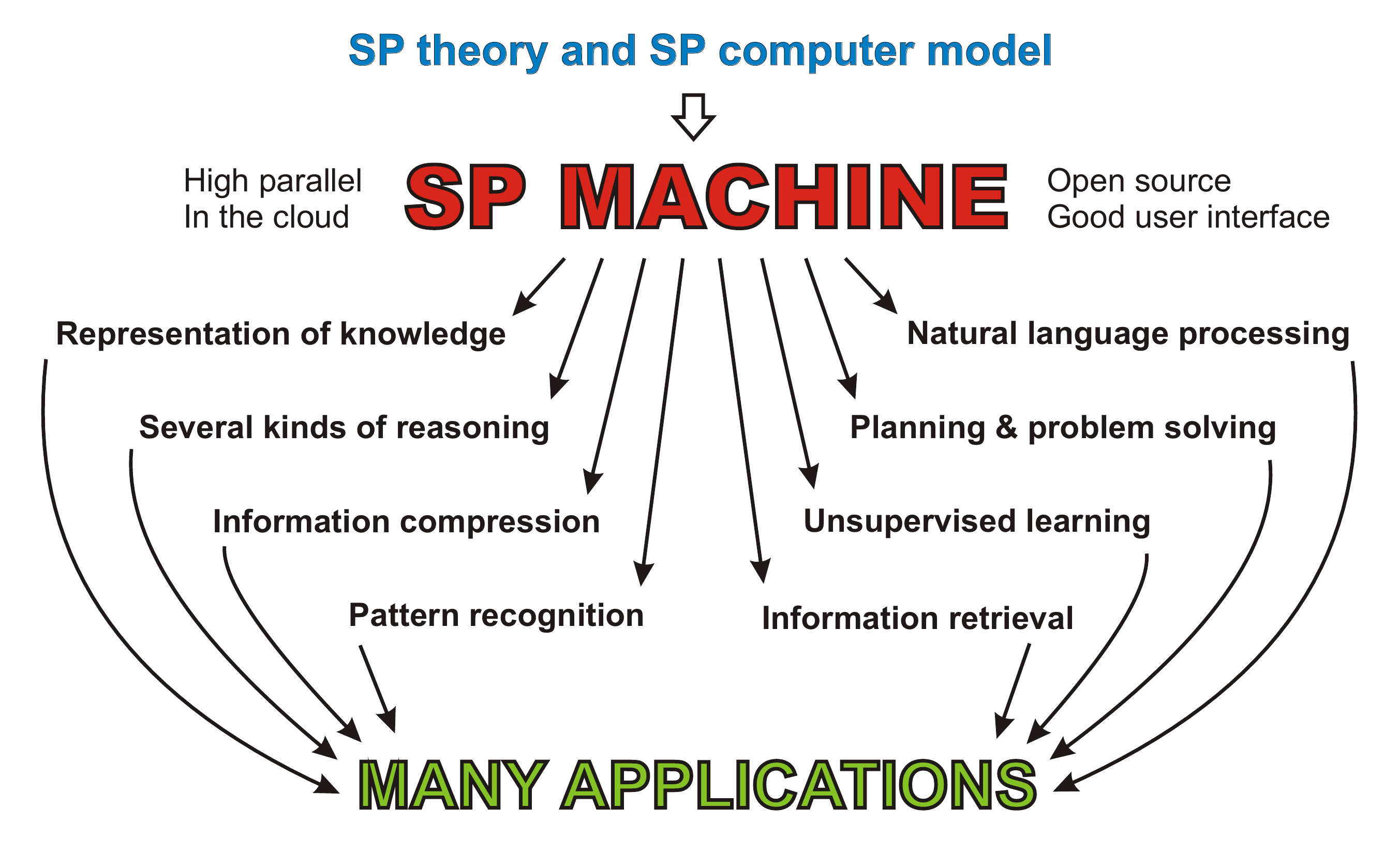}
\caption{Schematic representation of the development and application of the SP machine.}
\label{sp_machine_figure}
\end{figure}

As mentioned in the Introduction, it is intended that SPF will facilitate the development of a high-parallel version of the SP machine, realised as an open-source software virtual machine, and hosted on an existing high-performance computer. How things may develop is shown schematically in Figure \ref{sp_machine_figure}.

Aspects of the proposal are described in the following subsections.

\subsection{The proposed facility and public policies}\label{spf_and_public_policies}

Some aspects of SPF may seem unfamiliar and may, superficially, appear to conflict with funding policies. These remarks apply, in particular, to the proposal that SPF should be permanent (Section \ref{spf_permanent_section}), that it should be available to researchers anywhere in the world (Section \ref{open_to_researchers_everywhere_section}), and that, with few if any exceptions, there should be no charges for using the facility (Section \ref{easy_to_access_and_to_use_section}).

\sloppy But it appears that, at high levels, it is recognised that funding for e-infrastructure should be responsive and flexible, that it should recognise the diverse needs of the research community, and that there is a need for long-term perspectives. In support of such policies:

\begin{itemize}

\item ``In the roadmap, EPSRC aims to ...~understand the requirements of the EPS [engineering and physical sciences] research community that make use of e-infrastructure, ensuring there are no gaps or duplication.'' (From p.~3 of ``EPSRC e-infrastructure roadmap'', Engineering and Physical Sciences Research Council, August 2014, \href{http://bit.ly/1HvM34x}{bit.ly/1HvM34x}).

\item ``The funding agencies should recognise the need for flexibility of funding and put in place mechanisms to enable rapid responses to hardware and software developments and to new research opportunities and requirements.'' (Recommendation \#3 from p.~17 in ``Strategy for the UK Research Computing Ecosystem'', The e-Science Institute at The University of Edinburgh, October 2011, \href{http://bit.ly/1KslhrQ}{bit.ly/1KslhrQ}).

\end{itemize}

\subsection{Elements of SPF}\label{elements_of_spf_section}

It is envisaged that the main features of SPF will be these:

\begin{itemize}

\item {\em The SP machine, version 1}. SPF will contain the first high-parallel version of the SP machine, which may be referred to as the ``SP machine, version 1'' or ``SPM1''.\footnote{Although the SP computer model may be seen as a version of the SP machine (Appendix \ref{computer_model_and_sp_machine_appendix}), we think it is probably best if the first SPF version is called ``SPM1''.} The name ``SPM'' would normally mean SPM1 but, depending on the context, it may be used to refer to any other version of the SP machine or to all versions, collectively.

\item {\em Founded on the SP71 computer model}. The SP71 computer model (Appendix \ref{computer_model_and_sp_machine_appendix}), the basis for SPM, will be ported on to the host machine. This should be relatively straightforward to do since the model is written in C++, with good comments, and since a slightly earlier version, SP70, is described quite fully, with pseudocode, in \cite[Sections 3.9, 3.10, and 9.2]{wolff_2006}.

\item {\em High levels of parallel processing}. As far as possible, processes in the SP71 code will be modified to take advantage of parallel processing. This may be done in three main parts of the system, in procedures for: the matching and unification of patterns (as outlined in Appendix \ref{mup_appendix}); the building of multiple alignments (Appendix \ref{ma_appendix}); and the unsupervised learning of grammars (Appendix \ref{usl_grammars_appendix}).

\item {\em Good user interface}. SPF will be accessed via a well-designed website, with user-friendly interfaces for each part of the system, using multimedia where appropriate. It would be best if users did not have to download and install any software. Any such downloading of software can probably be avoided by developing the user interface using the facilities of HTML5. In accordance with Google's new search policies, announced in April 2015,\footnote{See, for example, ``Google search changes will promote mobile-friendly sites'', {\em BBC News}, 2015-04-20, \href{http://bbc.in/1Q9AORA}{bbc.in/1Q9AORA}.} the website should be ``mobile friendly'', meaning that it should be easy to use and legible on mobile phones.

\item {\em Setting up accounts}. There will be procedures for the registration of each user, assigning a unique identifier and setting up facilities for each account.\footnote{As noted in Section \ref{impact_section}, each user would be asked to ensure that, in all academic publications arising from research with SPF, there would be an acknowledgement of the facility and of the body or bodies that have provided the necessary funds.} Here, a ``user'' will be a research group or an individual researcher. To avoid undue complications, at least initially, each research group will be treated as if it was an individual researcher, with no attempt to differentiate individual researchers within each group, or their work. Each research group would need to make its own arrangements for the division of labour within the group.

    It is likely that most of the necessary mechanisms are already established for registering users of the host machine.

\item {\em Creating new versions of the SP machine}. Any user may create one or more new versions of SPM:

\begin{itemize}

\item Each such version will have its own storage space with documentation, source files, executable files, data files, and so on.

\item There will an hierarchical scheme of the usual kind for the identifiers of different versions, with identifier(s) for the creator(s) of each version.

\item The documentation should make clear the origins of each version in one or more previous versions, which may include SPM1 or other versions by the same user or other users.

\item Any version of SPM will be in one of two states: ``under development'' or ``completed''. Once a version has been ``completed'', further modifications will not be allowed. The executable code of any completed version may be run by any other user (next).

\end{itemize}

\item {\em Running the SP machine}. Any version of SPM that is under development may be run by the user that is developing it. Any completed version of SPM may be run by any user, which would of course include the user that created it. For any user to run any given version of SPM, facilities will be needed for:

\begin{itemize}

\item Assigning an area of storage to hold data and results.

\item Setting parameters for the given version of SPM.

\item Creating and editing SP patterns.

\item Uploading SP patterns that have been created elsewhere.

\item Online debugging of software (for SPM versions that are under development). It is likely that a suitable debugger is already established on the host machine.

\item Viewing results, including multiple alignments and grammars created by the program. Since some results---multiple alignments in particular---may be larger than can be seen comfortably on a typical computer screen, there should be facilities for zooming and scrolling.

\item Downloading SP patterns, included those newly-created by the given version of SPM.

\end{itemize}

\item {\em SP-neural, version 1}. Alongside SPM1 will be a first, tentative, version of the SP-neural machine, which may be referred to as ``SPNRL1''. The main elements of the SP-neural facility will be:

\begin{itemize}

\item As with SPM, each user may create one or more versions of SPNRL.

\item SPNRL1 will have a means of translating a set of SP patterns (or one or more multiple alignments) into an inter-connected set of pattern assemblies. With the object-oriented facilities of C++, there will be a class for neurons and a class for pattern assemblies.

\item In this preliminary model, each SP symbol will be represented by a single neuron although it is likely that in living brains, neural ``symbols'' are realised with clusters of inter-connected neurons.

\item Connections between neurons will be represented with C++ pointers. As SPNRL is developed, it is likely that there will be excitatory connections and also inhibitory connections, in accordance with what is known about neural tissue. It is also likely that there will be lateral connections between neighbouring neurons within each pattern assembly.

\item There will be a means of displaying, in graphical form, a set of inter-connected neurons and pattern assemblies, much as in Figure \ref{connections_figure}.

\item For any given set of inter-connected neurons and pattern assemblies, there will be a means of downloading it as a set of SP patterns.

\item As SPNRL is developed:

 \begin{itemize}

 \item There will be a means of sending excitatory and inhibitory signals through inter-connected pattern assemblies and displaying the results in terms of the excitatory levels of individual neurons.

 \item There will a means of creating new pattern assemblies and learning neural `grammars', in accordance with learning processes in any given version of SPM.

\item Parallel processing will be applied to mimic the high levels of parallelism that exist in the workings of real neural tissue.

\end{itemize}

\end{itemize}

\end{itemize}

\subsection{The facility should be permanent}\label{spf_permanent_section}

For reasons that follow, SPF should be permanent in the sense that it should be available for the foreseeable future:

\begin{itemize}

\item {\em Bridging the `Valley of Death'}. In accordance with ``It is our historic failure to back [R\&D, technology and engineering] which lies behind the familiar problems of the so-called `valley of death' between scientific discoveries and commercial applications'',\footnote{Speech by the Rt Hon David Willetts MP, ``The `eight great technologies' which will propel the UK to future growth receive a funding boost'', 2013-01-24, \href{http://bit.ly/1wKCqII}{bit.ly/1wKCqII}.} the facility should support and encourage research that has sufficient breadth and depth to bridge that `valley of death'. Short-termism is just as damaging for research as it is for business.

\item {\em Long-term research programmes}:

\begin{itemize}

\item {\em Researchers need confidence that SPF will not be withdrawn}. We envisage the proposed facility providing the basis for research that is likely to include long-term research programmes. Any research group that wishes to embark on that kind of research programme needs to have confidence that the research facility will not be withdrawn. {\em Without assurance for researchers that the facility is permanent, it is likely that the entire project will fail}.

\item {\em BIS: software as a sustained infrastructure in the long term}. ``Current investments in software development should be reviewed with a view to developing models of support for software as a sustained infrastructure in the long term, as opposed to being supported by significant one off investments.''\footnote{From p.~8 in ``Report of the e-Infrastructure Advisory Group'', Department for Business, Innovation and Skills, June 2011, \href{http://bit.ly/1LMXOSD}{bit.ly/1LMXOSD}.}

\item {\em e-Science Institute: the need for long-term funding for ambitious software development projects}. ``There needs to be long-term funding for ambitious software development projects .... The current approach is hardware centric and short-term ..., but does not recognise the critical and long-term role that software and people play in the [research computing] enterprise.''\footnote{Recommendations \#4 and \#5, p.~17, in ``Strategy for the UK Research Computing Ecosystem'', The e-Science Institute at The University of Edinburgh, October 2011, \href{http://bit.ly/1KslhrQ}{bit.ly/1KslhrQ}.}

\end{itemize}

\item {\em Time to develop research programmes}. It is likely to take time for a research group to learn about the SP system, to decide to mount a research project, to prepare a research proposal, for the proposal to be assessed, to provide PCs and other facilities, and to recruit staff. If the proposed research facility is only available for a few years (bearing in mind the time needed to create the facility), much of that time may be taken up with the processes described. Then it is likely that there would be little or no time left to do the actual research.

\item {\em Consistency across facilities}. In the same way that large and expensive facilities---large telescopes, atom smashers, supercomputers, and the like---are maintained for many years, and replaced as necessary, the same should apply to less expensive or smaller-scale facilities. In both cases, research benefits from the long-term perspective that stability makes possible.

\item {\em Archival status}. As outlined in Section \ref{elements_of_spf_section}, the facility will be designed so that research groups and individual researchers may create their own versions of the SP machine. Each such version needs to have archival status:

\begin{itemize}

\item {\em The need for long-term access to and execution of software}. Other researchers, now and in the future, need to be able to access the source code and documentation for any version of the SP machine and to run the executable code. Naturally, the host machine for SPF may need to be replaced from time to time. In any such case, SPF, including all associated projects and data, may be ported on to the replacement machine, in accordance with standard practice in industry and commerce.

\item \sloppy {\em RCUK: software as a vital research infrastructure}. The need for long-term preservation of software and associated computing systems is implied by:``Software developed for experimental facilities and instrumentation, modelling and simulation and data-analysis is a critical and valuable resource. Software and algorithm development represents major investments by skilled researchers, and the large suite of codes and algorithms used in research should be regarded as a vital research infrastructure, requiring support and maintenance along the innovation chain, and throughout its life cycle. The reproducibility of research is at the very heart of the scientific method.''\footnote{From p.~8 of ``E-infrastructure roadmap'', RCUK e-Infrastructure Group, 2014, \href{http://bit.ly/1Jx59EE}{bit.ly/1Jx59EE}.}.

\item {\em Long-term preservation of software}. This requirement accords with the trend for academic journals to allow the author or authors of any published paper to provide supplementary material, which may include data, software, and things that may be regarded as forms of software, such as MP4 files.\footnote{The journal {\em Royal Society Open Science} says ``Supplementary material can be used for supporting data sets, supporting movies, figures and tables, and any other supporting material.'' (``Instructions for Authors'', \href{http://bit.ly/1wD8ezN}{bit.ly/1wD8ezN}); and {\em IEEE Access} allows authors to include ``data collections and multimedia materials'' (``IEEE Access---Information for Authors'', PDF, \href{http://bit.ly/1wD5H8x}{bit.ly/1wD5H8x}).} It is increasingly recognised that many bodies of software should have the same archival status as peer-reviewed papers in academic journals or conference proceedings, and bodies of data on which those papers are based.

\item {\em Preserving the execution environment}. Since software will often not run, or will not run correctly, unless it is running on the type of computer (with its operating system and related software) for which it was written, there is also a need for long-term preservation of each type of computer. There is more detail here:

\begin{itemize}

\item {\em Vint Cerf: digital ``dark age'' and ``digital vellum''}. The case has been argued by internet pioneer Vint Cerf, as reported in, for example: ``Google's Vint Cerf warns of `digital Dark Age'\thinspace''\footnote{BBC News, 2015-02-13, \href{http://bbc.in/1D3pemp}{bbc.in/1D3pemp}}. Vint Cert also makes the case in a lecture that he gave on 2015-02-11, ``Digital vellum and the expansion of the internet into the solar system'' (see \href{http://bit.ly/1aqDeKw}{bit.ly/1aqDeKw});

\item \sloppy {\em Olive Executable Archive}. How software and associated computing systems may be preserved is demonstrated in the ``Olive Executable Archive'', \href{https://olivearchive.org/}{www.olivearchive.org}. This project has been highlighted by Vint Cerf in his articles and lectures.

\item {\em New gold standard established for open and reproducible research}. The case has also been argued by computer scientists at Cambridge University who ``have set a new gold standard for openness and reproducibility in research by sharing the more than 200GB of data and 20,000 lines of code behind their latest results---an unprecedented degree of openness in a peer-reviewed publication.''\footnote{``New gold standard established for open and reproducible research'', Cambridge University press release, 2015-05-04, \href{http://bit.ly/1IgFkKG}{bit.ly/1IgFkKG}.}

\item {\em Maintaining access to digitised information}. The need to maintain access to digitised information in the face of changes in the hardware or software environment that it needs is discussed in ``Preserving the digital record of computing history''.\footnote{Article by David Anderson in {\em Communications of the ACM}, vol.~58, no.~7, July 2015, pp.~29--31.}

\end{itemize}

    As noted above, replacement of SPF's host machine should not be a problem. The system may be ported onto any new machine, as or when that becomes necessary.

\end{itemize}

\item {\em Avoiding waste and gaining full value from SPF}. Given that the bulk of the cost of SPF is in its initial development and setting up, there is little scope for saving money by terminating the facility within a few years. Indeed, any such move is likely to mean a waste of resources and a failure to gain full value from the facility:

\begin{itemize}

\item {\em The damaging effect of short-termism}. As noted above, it is likely that the entire project will fail unless researchers have confidence that the facility is permanent.

\item {\em Loss of valuable research}. Premature termination of SPF would mean losses of valuable research that may otherwise accrue, much of it provided at no cost to the UK by researchers elsewhere (Sections \ref{impact_section} and \ref{easy_to_access_and_to_use_section}).

\item {\em SPF will not wear out and its value will increase}. Unlike ordinary research facilities such as space telescopes or atom smashers, SPF will not wear out. And its value for researchers will increase progressively as new versions of SPM are created and made available for the whole research community, including all new insights and improvements.

\end{itemize}

\end{itemize}

\subsection{The facility should be available to researchers anywhere in the world}\label{open_to_researchers_everywhere_section}

As previously noted, the wide scope of the SP system means that there are far more avenues to be explored than any one research group could tackle on its own. Accordingly, we are are aiming to create a facility for research groups and individual researchers anywhere in the world. It should not be restricted to UK scientists:

\begin{itemize}

\item This accords with what John Kelly and Steve Hamm say about the need for large-scale collaboration (quoted in Section \ref{theory_and_machine_section}) and also with the need for integration of e-infrastructures ``internationally, across other national e-infrastructures, to deliver end-to-end services in the global environment of collaborative research''\footnote{p.~2 in {\em E-infrastructure roadmap}, Research Councils UK, \href{http://bit.ly/1Jx59EE}{bit.ly/1Jx59EE}.} and the need to facilitate ``... research collaboration between industry and academia.''\footnote{{\em ibid.}, p.~5.}

\item It also accords with the principles that:

\begin{itemize}

\item The EPSRC aims to ``Understand the whole UK e-infrastructure landscape, view it holistically and {\em consider it within an international context} (emphasis added).''\footnote{The EPSRC ``E-Infrastructure Roadmap'', retrieved 2015-07-01, \href{http://bit.ly/1R3V767}{bit.ly/1R3V767}.}

\item ``The UK’s Research and Innovation e-infrastructure needs to be led and driven to deliver a UK wide vision for research e-infrastructure, {\em embedded in the international context essential to today’s research challenges}.'' (emphasis added).\footnote{``Report of the e-Infrastructure Advisory Group'', Department of Business, Innovation and Skills, June 2011, p.~3, \href{http://bit.ly/1LMXOSD}{bit.ly/1LMXOSD}.}

\end{itemize}

\item Opening up the research in this way is likely to yield synergies that might not arise if there were more restrictions (Section \ref{uk_industrial_strategy_section})---with corresponding benefits for UK researchers.

\item The international perspective will help to achieve full value from the facility (Section \ref{broad_scope_section}).

\item And it will help to increase the impact of the facility (Section \ref{impact_section}) and its international influence (Section \ref{international_influence_section}).

\end{itemize}

\subsection{The facility should be easy to access and easy to use}\label{easy_to_access_and_to_use_section}

In keeping with the main points in the preceding subsection, it is important to minimise obstacles to the use of the system for preliminary investigations, for teaching students, for individual research projects, and for long-term research programmes. An additional reason is that, for any research group or individual researcher that is not already familiar with the SP concepts, a significant amount of time and effort is likely to be required to get up to speed, so it is important to minimise other obstacles to progress.

For these reasons:

\begin{itemize}

\item {\em Ease of access}. The process of registering with the system and gaining access to it should be as simple and straightforward as possible.

\item {\em User friendly}. For all aspects and parts of the system, there should be a user-friendly interface.

\item {\em Free at the point of use}. With few if any exceptions, the system should be free for users. In addition to minimising obstacles for researchers, reasons include:

\begin{itemize}

\item {\em Research dividends}. In keeping with remarks about research `dividend' in Section \ref{impact_section}, the overall value of research done with SPF is likely to outweigh the cost of providing and maintaining the facility, and by a considerable margin.

\item {\em Research bonuses}. As noted in Section \ref{impact_section}, research that is done by research groups and individual researchers outside the UK may be seen as a bonus, additional to what would be achieved if the facility were restricted to UK researchers.

\item {\em Research synergies}. Also, as noted in Section \ref{uk_industrial_strategy_section}, there are likely to be benefits from synergies and large-scale collaboration.

\item {The `Valley of Death'}. In accordance with remarks in Section \ref{spf_permanent_section} about the `valley of death', it is important to avoid any premature requirement that the research should, commercially, ``stand on its own two feet'' or ``become self-sustaining'', in much the same way that we do not expect young children to go out and earn a living.

\end{itemize}

\end{itemize}

With regard to costs, these may be met as follows:

\begin{itemize}

\item {\em Development and maintenance costs}. These costs may be met from a variety of sources, perhaps including funds for the support of researchers based in the UK.

\item {\em Running costs}. Although it is likely that mature versions of SPM will be used for computationally-demanding tasks, we believe that much research with SPF can be done with small examples that make small computational demands on the host computer, and that the associated costs would be relatively small. It seems likely that the cost of administering a system of charges for running costs would outweigh the actual costs. In that case, it would be best, probably, not to attempt to make any charges for the use of the system. A possible exception is where a user wishes to run the system with computational tasks that are likely to consume large computational resources.

\end{itemize}

\subsection{Open source}

As already mentioned, SPM should be open source. More specifically, it should be possible for anyone---including people who are not registered users of SPF---to view the source code and documentation of any completed version of SPM, and for any registered user to run any completed version.

\sloppy In keeping with these points, all versions of SPM should be ``free software'' conforming to the GNU General Public License, described on \href{http://www.gnu.org/licenses/gpl.html}{www.gnu.org/licenses/gpl.html}. In brief, this means that every user should have:

\begin{itemize}

\item The freedom to use the software for any purpose;

\item The freedom to change the software to suit his or her needs;

\item The freedom to share the software with friends and neighbours; and

\item The freedom to share the changes that any user makes.

\end{itemize}

This policy conforms to and supports the long-established principle that science works best when ideas, observations and experimental results are freely available to everyone, without restrictions. This kind of openness and transparency is essential if we are to make the difficult transition to cognitive computing (Section \ref{challenges_section}).

\section{Costs}\label{costs_section}

Costs associated with SPF are estimated in the following subsections. A spreadsheet that summarises all the estimated costs, with totals, is shown in Section \ref{summary_of_costs_section}.

\subsection{Development costs}\label{development_costs_section}

We estimate that three members of staff with the right skills who are already familiar with the host machine would be able to develop SPF within 2 years. We suggest that it would be appropriate to employ one relatively senior person, perhaps at a salary of about $\pounds60,000$, one more junior person, perhaps at a salary of about $\pounds40,000$, and a third person, possibly a student, at a salary of about $\pounds20,000$.

We have assumed that the cost of employing someone is twice the cost of his or her salary, and of course salaries would be paid for two years.

We have allowed $\pounds 15,000$ for PCs, printers, software, and related facilities for the three staff.

In view of the importance of the website for SPF and the specialised skills required for good website design---including the skills needed to ensure that the website is ``mobile friendly''---we believe it would be prudent to employ a web design company to establish the framework for the website and advise on how it should be developed. We estimate that $\pounds 2000$ should be allowed for this work.

We have allowed 10\% for contingencies.

\subsection{Recurrent costs}

The annual cost of running SPF are estimated in the following subsections and summarised in Section \ref{summary_of_costs_section}.

\subsubsection{Maintenance costs}\label{maintenance_costs_section}

We believe that when SPF is up and running, it will need support: to make sure that it runs smoothly and to deal with any snags that arise; to answer queries by users; to fix bugs in the system; and to refine the system in the light of feedback from users.

We believe that these things can be done by one person with a salary of about $\pounds40,000$ pa, with an assistant at about $\pounds20,000$ pa.

\subsubsection{Running costs}\label{running_costs_section}

As mentioned in Section \ref{easy_to_access_and_to_use_section}, we believe that much of the research with SPF would make small demands on the host computer and, probably, that the associated costs would be smaller than the cost of administering a system of charges. Nevertheless, such costs do need to be covered. We estimate that they would be $\pounds 10,000$ pa.

\subsubsection{Awareness-raising costs}\label{awareness-raising_costs_section}

In accordance with ``impact'' policies of the UK Government and UK Research Councils (Section \ref{impact_section}), it is important that SPF and what it can do should be well known around the world, amongst academic and industrial researchers, amongst research administrators and policy-makers, and amongst members of the general public. This may be achieved via the following main routes, each one with associated costs:

\begin{itemize}

\item Via papers in academic journals and conference proceedings. Most of these would be prepared by research groups and individual researchers that are using SPF, and they would pay the associated costs. But, in addition, there would be academic papers about SPF and SPM, prepared by the two authors of this proposal and staff recruited to develop SPF. The main costs here would be the costs of attending conferences and charges for open-access publication.

\item The proposers, and perhaps development staff as well, may travel to give talks about SPF and the SP concepts.

\item Perhaps via conferences (online or traditional) dedicated to research using SPF. The organisation may be undertaken by a specialist conference organiser, and there would be associated costs.

\item Via articles in popular-science magazines, both those for the general reader and those with a more technical or specialist orientation. Here, it would be useful if articles could be prepared by writers with relevant skills (who would need to be paid), although technical input would be provided by the authors of this proposal and research/development staff. Articles may be translated from English into other languages.

\item Via the mainstream media (newspapers, radio and TV), internet blogs, and online videos. Again, it would be useful if articles, videos and the like could be prepared by people with relevant skills, with guidance from people with relevant technical knowledge.

\end{itemize}

We estimate that a budget of $\pounds 50,000$ pa would be needed to cover these costs during the two-year development phase, reducing to $\pounds 20,000$ pa for a further three years. Although we believe it is important that SPF should be a permanent facility, as described in Section \ref{spf_permanent_section}, we shall assume for present purposes that awareness-raising costs will be spread over a period of 20 years. In that case, the annual cost will be $\pounds ((50,000 \times 2) + (20,000 \times 3)) / 20 = \pounds 8,000$.

\subsubsection{Contingencies}

As with the non-recurrent development costs, we have allowed 10\% for contingencies.

\subsection{Summary of costs}\label{summary_of_costs_section}

In this section, Table \ref{spreadsheet_table} summarises our estimates of non-recurrent costs for SPF ($\pounds 546,700$ in total) and recurrent costs ($\pounds 151,800$ pa in total).

\begin{table}[!htbp]
\begin{center}
\begin{tabular}{l r r}
\em Category & \em $\pounds$ & \em $\pounds$ \\
\\
{\bf Non-recurrent (development)} \\

Development salaries \\

~~~Senior & 70,000 \\

~~~Middle & 45,000 \\

~~~Junior & 25,000 \\

Total (salaries, one year) & 140,000 \\

Total (salaries $+$ overheads, one year) & 280,000 \\

Total (salaries $+$ overheads, two years) & & 560,000 \\

PCs etc & & 15,000 \\

Website design & & 2,000 \\

Sub-total (non-recurrent) & 577,000 \\

Contingencies (10\%) & & 57,700 \\

{\bf Total (non-recurrent)} & & 634,700 \\

\\

{\bf Recurrent (per year)} \\

Maintenance salaries \\

~~~Middle & 45,000 \\

~~~Junior & 25,000 \\

Total (salaries) & 70,000 \\

Total (salaries $+$ overheads) & & 140,000 \\

Running costs & & 10,000 \\

Awareness raising & & 8,000 \\

Sub-total (recurrent) & 158,000 \\

Contingencies (10\%) & & 15,800 \\

{\bf Total (recurrent)} & & 173,800 \\

\end{tabular}
\end{center}
\caption{A spreadsheet summarising our estimates of non-recurrent and recurrent costs for SPF.}
\label{spreadsheet_table}
\end{table}

\section{Conclusion}\label{conclusion_section}

The wide scope of the SP theory and the wide range of its potential benefits and applications means that there are many more avenues to be explored than could be tackled by any one research group. By providing the means for researchers worldwide to participate and collaborate, the research facility that is proposed in this document would facilitate the further development of the SP system and the realisation of its potential.

In view of that potential, and the relatively small cost of the proposed facility, the creation of that facility would be a good investment.

% \newpage

\section*{Appendices}
\addcontentsline{toc}{section}{Appendices}

\appendix

\section{Outline of the SP theory and SP computer model}\label{outline_of_sp_theory_and_model_appendix}

\begin{figure}[!htbp]
\centering
\includegraphics[width=0.5\textwidth]{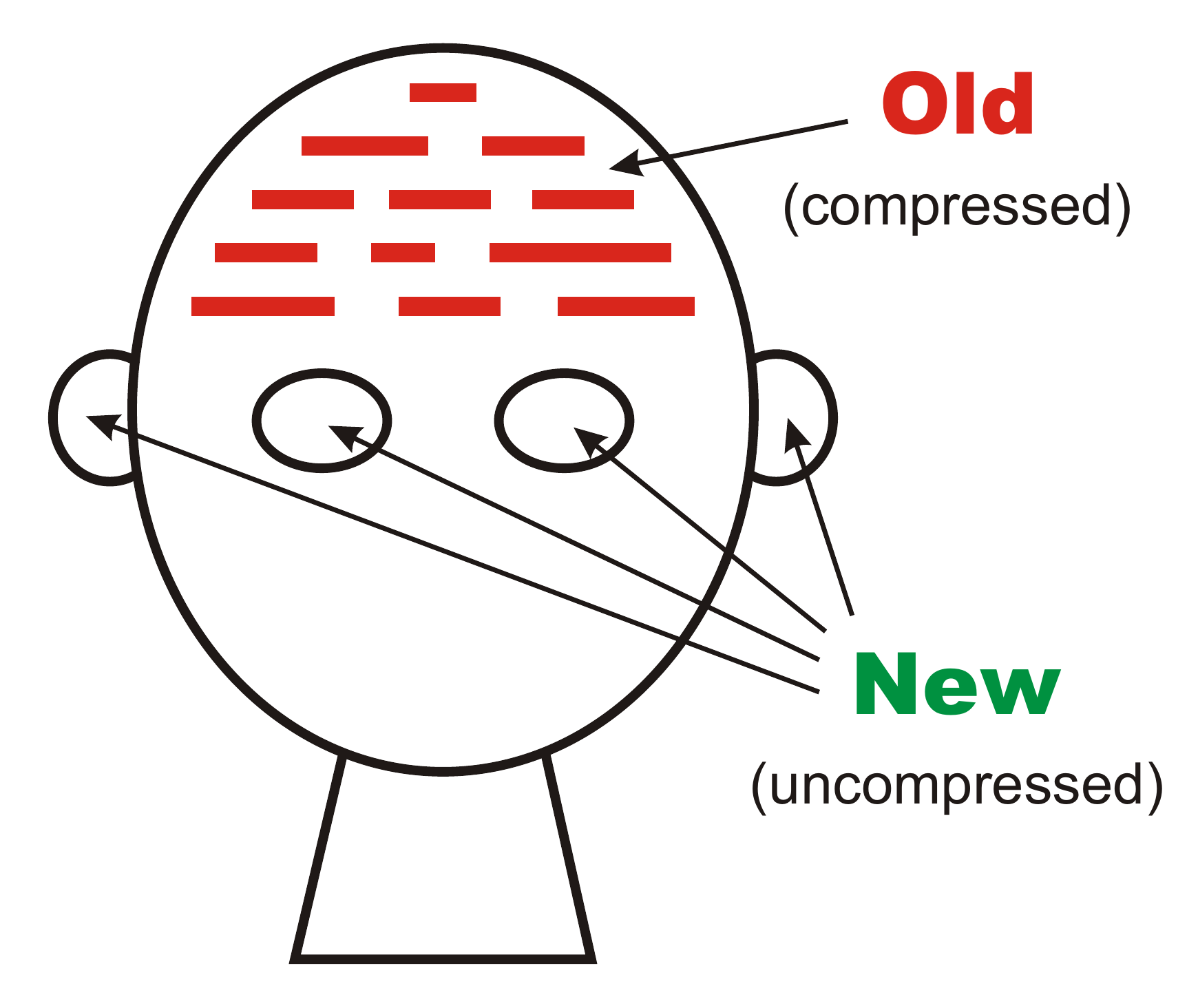}
\caption{Schematic representation of the SP system from an `input' perspective.}
\label{sp_input_perspective_figure}
\end{figure}

The SP theory is conceived as an abstract brain-like system that, in an `input' perspective, may receive {\em New} information via its senses, and compress some or all of it to create {\em Old} information, as illustrated schematically in Figure \ref{sp_input_perspective_figure}. In the theory, information compression is the mechanism both for the learning and organisation of knowledge and for pattern recognition, reasoning, problem solving, and more.

The subsections that follow outline the main elements of the SP theory and the SP machine.

\subsection{The SP computer model and the SP machine}\label{computer_model_and_sp_machine_appendix}

The SP theory is realised in the form of a computer model, SP71, which may be regarded as a version of the {\em SP machine}.

As noted in the Introduction, the use of a computer model as a vehicle for the theory has helped to reduce vagueness in the theory, it has provided a means of testing candidate ideas, many of which have been rejected, and it is a means of demonstrating what can be done with the system.

An outline of how the SP computer model works may be found in \cite[Section 3.9]{wolff_2006}, with more detail, including pseudocode, in \cite[Sections 3.10 and 9.2]{wolff_2006}.\footnote{These sources describe SP70, a slightly earlier version of the model than SP71 but quite similar to it. The description of SP70 includes a description, in \cite[Sections 3.9.1 and 3.10]{wolff_2006}, of a subset of the SP70 model called SP61.} Fully commented source code for the SP71 computer model may be downloaded via a link near the bottom of \href{http://www.cognitionresearch.org/sp.htm}{www.cognitionresearch.org/sp.htm}, and via ``Ancillary files'' under \href{http://arxiv.org/abs/1306.3888}{www.arxiv.org/abs/1306.3888}.

As previously noted, we envisage that the SP computer model will be the basis for the creation of the high-parallel, open-source version of the SP machine. How things may develop is shown schematically in Figure \ref{sp_machine_figure}.

\subsection{Patterns and symbols}\label{patterns_and_symbols_appendix}

In the SP system, knowledge is represented with arrays of atomic symbols in one or two dimensions called {\em patterns}. The SP71 model works with 1D patterns but it is envisaged that the system will be generalised to work with 2D patterns \cite[Section 3.3]{sp_extended_overview}.

An `atomic symbol' in the SP system is simply a mark that can be matched with any other symbol to determine whether it is the same or different: no other result is permitted.

In themselves, SP patterns are not particularly expressive. But within the multiple alignment framework (Appendix \ref{ma_appendix}), they support the representation and processing of a wide variety of kinds of knowledge (Appendix \ref{ma_knowledge_appendix}). A goal of the SP research programme is to establish one system for the representation and processing of {\em all} kinds of knowledge (see also \cite[Section III]{sp_big_data}). Evidence to date suggests that this may be achieved with SP patterns in the multiple alignment framework.

Any collection of SP patterns is termed a {\em grammar}. Although that term is most closely associated with linguistics, it is used in the SP research programme for a collection of SP patterns describing any kind of knowledge.

\subsection{Information compression}\label{information_compression_appendix}

In the SP theory, the emphasis on information compression derives from earlier research on grammatical inference \cite{wolff_1988} and the principle of {\em minimum length encoding} (MLE) \cite{solomonoff_1964,wallace_boulton_1968,rissanen_1978}).

At an abstract level, information compression means the detection and reduction of {\em redundancy} in information. In more concrete terms, redundancy means {\em recurrent patterns}, {\em regularities}, {\em structures}, and {\em associations}, including {\em causal associations}. Thus information compression provides a means of discovering such things as words in natural language \cite{wolff_1988}, objects \cite[Section V-E]{sp_autonomous_robots}, and associations (see, for example, \cite[Section III-A.1]{sp_big_data}), in accordance with the DONSVIC principle \cite[Section 5.2]{sp_extended_overview}.\footnote{{\em DONSVIC} = ``The discovery of natural structures via information compression''.}

The default assumption in the SP theory is that compression of information is always lossless, meaning that all non-redundant information is retained. In particular applications, there may be a case for discarding non-redundant information (see, for example, \cite[Section X-B]{sp_big_data}) but any such discard is reversible.

In the SP system, information compression is achieved via the matching and unification of patterns. More specifically, it is achieved via the building of multiple alignments and via the unsupervised learning of grammars. These three things are described briefly in the following three subsections.

\subsubsection{Information compression via the matching and unification of patterns}\label{mup_appendix}

The basis for information compression in the SP system is a process of searching for patterns that match each other with a process of merging or `unifying' patterns that are the same. At the heart of the SP71 model is a method for finding good full and partial matches between sequences with advantages compared with classical methods \cite[Appendix A]{wolff_2006}.\footnote{The main advantages are \cite[Section 3.10.3.1]{wolff_2006}: 1) That it can match arbitrarily long sequences without excessive demands on memory; 2) For any two sequences, it can find a set of alternative matches (each with a measure of how good it is) instead of a single `best' match; 3) The `depth' or thoroughness of the searching can be controlled by parameters.}

\subsubsection{Information compression via the building of multiple alignments}\label{ma_appendix}

That process for finding good full and partial matches between patterns is the foundation for processes that build {\em multiple alignments} like the one shown in Figure \ref{class_part_plant_figure}. This concept is similar to multiple alignment in bioinformatics but with important differences \cite[Section 3.4]{wolff_2006}. It is a powerful and distinctive feature of the SP system.

\newgeometry{margin=2cm,includefoot}

\begin{figure}[!htbp]
\fontsize{07.00pt}{08.40pt}
\centering
{\bf
\begin{BVerbatim}
0                 1                2                  3              4              5                  6

                  <species>
                  acris
                  <genus> ---------------------------------------------------------------------------- <genus>
                  Ranunculus ------------------------------------------------------------------------- Ranunculus
                                                                                    <family> --------- <family>
                                                                                    Ranunculaceae ---- Ranunculaceae
                                                                     <order> ------ <order>
                                                                     Ranunculales - Ranunculales
                                                      <class> ------ <class>
                                                      Angiospermae - Angiospermae
                                   <phylum> --------- <phylum>
                                   Plants ----------- Plants
                                   <feeding>
has_chlorophyll ------------------ has_chlorophyll
                                   photosynthesises
                                   <feeding>
                                   <structure> ------ <structure>
                                                      <shoot>
<stem> ---------- <stem> ---------------------------- <stem>
hairy ----------- hairy
</stem> --------- </stem> --------------------------- </stem>
                  <leaves> -------------------------- <leaves>
                  compound
                  palmately_cut
                  </leaves> ------------------------- </leaves>
                                                      <flowers> ------------------- <flowers>
                                                                                    <arrangement>
                                                                                    regular
                                                                                    all_parts_free
                                                                                    </arrangement>
                  <sepals> -------------------------------------------------------- <sepals>
                  not_reflexed
                  </sepals> ------------------------------------------------------- </sepals>
<petals> -------- <petals> -------------------------------------------------------- <petals> --------- <petals>
                                                                                    <number> --------- <number>
                                                                                                       five
                                                                                    </number> -------- </number>
                  <colour> -------------------------------------------------------- <colour>
yellow ---------- yellow
                  </colour> ------------------------------------------------------- </colour>
</petals> ------- </petals> ------------------------------------------------------- </petals> -------- </petals>
                                                                                    <hermaphrodite>
<stamens> ------------------------------------------------------------------------- <stamens>
numerous -------------------------------------------------------------------------- numerous
</stamens> ------------------------------------------------------------------------ </stamens>
                                                                                    <pistil>
                                                                                    ovary
                                                                                    style
                                                                                    stigma
                                                                                    </pistil>
                                                                                    </hermaphrodite>
                                                      </flowers> ------------------ </flowers>
                                                      </shoot>
                                                      <root>
                                                      </root>
                                   </structure> ----- </structure>
<habitat> ------- <habitat> ------ <habitat>
meadows --------- meadows
</habitat> ------ </habitat> ----- </habitat>
                  <common_name> -- <common_name>
                  Meadow
                  Buttercup
                  </common_name> - </common_name>
                                   <food_value> ----------------------------------- <food_value>
                                                                                    poisonous
                                   </food_value> ---------------------------------- </food_value>
                                   </phylum> -------- </phylum>
                                                      </class> ----- </class>
                                                                     </order> ----- </order>
                                                                                    </family> -------- </family>
                  </genus> --------------------------------------------------------------------------- </genus>
                  </species>

0                 1                2                  3              4              5                  6
\end{BVerbatim}
}
\caption{\small The best multiple alignment created by the SP computer model, with a set of New patterns (in column 0) that describe some features of an unknown plant, and a set of Old patterns, including those shown in columns 1 to 6, that describe different categories of plant, with their parts and sub-parts, and other attributes.}
\label{class_part_plant_figure}
\end{figure}

\restoregeometry

This example shows the best multiple alignment created by the SP computer model when a set of New patterns (in column 0)\footnote{Specifically, the New patterns in this example are `\texttt{has\_chlorophyll}' (a pattern with one symbol), `\texttt{<stem> hairy </stem>}', `\texttt{<petals> yellow </petals>}', `\texttt{<stamens> numerous </stamens>}', and `\texttt{<habitat> meadows </habitat>}'. The patterns in a set like that may be presented to the system and processed in any order.} is processed in conjunction with a set of pre-existing Old patterns like those shown in columns 1 to 6. Here, the multiple alignment is `best' because it is the one that achieves the most economical description of the New patterns in terms of the Old patterns. The way in which that description or `encoding' is derived from a multiple alignment is explained in \cite[Section 3.5]{wolff_2006} and \cite[Section 4.1]{sp_extended_overview}. Like all other kinds of knowledge, encodings derived from multiple alignments are recorded using SP patterns (Appendix \ref{patterns_and_symbols_appendix}).

This multiple alignment may be interpreted as the result of a process of recognition (Appendix \ref{pattern_recognition_appendix}). The New patterns represent the features of some unknown plant and the Old patterns in columns 1 to 6 show how the plant has been identified at several levels of abstraction: species `Meadow Buttercup' (column 1), genus {\em Ranunculus} (column 6), family {\em Ranunculaceae} (column 5), and so on.

\subsubsection{Information compression via the unsupervised learning of grammars}\label{usl_grammars_appendix}

As outlined in \cite[Section 3.9.2]{wolff_2006} and \cite[Section 5.1]{sp_extended_overview}, and described more fully in \cite[Chapter 9]{wolff_2006}, the SP system may, without assistance from a ``teacher'' or anything equivalent, derive one or more plausible grammars from a body of New patterns, with minimum length encoding as a guiding principle. In that process, multiple alignment has a central role as a source of SP patterns for possible inclusion in any grammar \cite[Section V-B1]{sp_autonomous_robots}.

\subsubsection{Heuristic search}\label{heuristic_search_appendix}

Like most problems in artificial intelligence, each of the afore-mentioned problems---finding good full and partial matches between patterns, finding or constructing good multiple alignments, and inferring one or more good grammars from a body of data---is normally too complex to be solved by exhaustive search.

With intractable problems like these, it is often assumed that the goal is to find theoretically ideal solutions. But with these and most other AI problems, ``The best is the enemy of the good''. By scaling back one's ambitions and searching for ``reasonably good'' solutions, it is often possible to find solutions that are useful, and without undue computational demands.

As with other AI applications, and as with the building of multiple alignments in bioinformatics, the SP71 model uses heuristic techniques in all three cases mentioned above. This means searching for solutions in stages, with a pruning of the search tree at every stage, guided by measures of success \cite[Appendix A; Sections 3.9 and 3.10; Chapter 9]{wolff_2006}. With these kinds of techniques, acceptably good approximate solutions can normally be found without excessive computational demands and with ``big O'' values that are within acceptable limits.

\subsection{Multiple alignment and the representation and processing of diverse kinds of knowledge}\label{ma_knowledge_appendix}

\sloppy The expressive power of SP patterns within the multiple alignment framework derives in large part from the way that symbols in one pattern may serve as links to one or more other patterns or parts thereof. One of several examples in Figure \ref{class_part_plant_figure} is how the pair of symbols `\texttt{<family>~...~</family>}' in column 6 serves to identify the pattern `\texttt{<family> ...~Ranunculales ...~<hermaphrodite> ...~poisonous ...~</family>}' in column 5.

In the figure, these kinds of linkages between patterns mean that the unknown plant (with characteristics shown in column 0) may be recognised at several different levels within a hierarchy of classes: genus, family, order, class, and so on. Although it is not shown in this example, the system also supports cross classification.

In the figure, the parts and sub-parts of the plant are shown in such structures as `\texttt{<shoot>}' (column 3), `\texttt{<flowers>}' (column 5), `\texttt{<petals>}' (column 6), and so on.

As in conventional systems for object-oriented design, the system provides for inheritance of attributes (Appendix \ref{reasoning_appendix}). But unlike such systems, there is smooth integration of class hierarchies and part-whole hierarchies, without awkward inconsistencies \cite[Section 4.2.1]{wolff_sp_intelligent_database}.

More generally, SP patterns within the multiple alignment framework provide for the representation and processing of a wide variety of kinds of knowledge including: the syntax and semantics of natural language; class hierarchies and part-whole hierarchies (as just described); networks and trees; entity-relationship structures; relational knowledge; rules and several kinds of reasoning; patterns and pattern recognition; images; structures in three dimensions; and procedural knowledge. There is a summary in \cite[Section III-B]{sp_big_data}, and more detail in Appendix \ref{potential_benefits_applications_appendix}.

\subsection{Information compression, prediction, and probabilities}\label{ic_prediction_probabilities_appendix}

Owing to the close connection between information compression and concepts of prediction and probability \cite{li_vitanyi_2009}, the SP system is fundamentally probabilistic.\footnote{There are reasons to believe that the same is true of ``computing'' at its most fundamental level, and also mathematics. For example, Gregory Chaitin has written ``I have recently been able to take a further step along the path laid out by G{\"o}del and Turing. By translating a particular computer program into an algebraic equation of a type that was familiar even to the ancient Greeks, I have shown that there is randomness in the branch of pure mathematics known as number theory. My work indicates that---to borrow Einstein’s metaphor---God sometimes plays dice with whole numbers.'' \cite[p.~80]{chaitin_1988}.} Each SP pattern has an associated frequency of occurrence and probabilities may be calculated for each multiple alignment and for any inference that may be drawn from any given multiple alignment.

Although the SP system is fundamentally probabilistic: it can be constrained to answer only those kinds of questions where probabilities are close to 0 or 1; and, via the use of error-reducing redundancy, it can deliver decisions with high levels of confidence. Contrary to what one may suppose, there is no conflict between the use of error-reducing redundancy and the notion that ``computing'' may be understood as information compression---the two things are independent, as described in \cite[Section 2.3.7]{wolff_2006}.

\subsection{SP-neural}\label{sp-neural_appendix}

Part of the SP theory is the idea, described most fully in \cite[Chapter 11]{wolff_2006}, that the abstract concepts of {\em symbol} and {\em pattern} in the SP theory may be realised more concretely in the brain with collections of neurons in the cerebral cortex.

The neural equivalent of an SP pattern is called a {\em pattern assembly}. The word ``assembly'' has been adopted in this term because the concept is quite similar to Donald Hebb's \cite{hebb_1949} concept of a {\em cell assembly}. The main difference is that the concept of pattern assembly is unambiguously explicit in proposing that the sharing of structure between two or more pattern assemblies is achieved by means of `references' from one structure to another, as described and discussed in \cite[Section 11.4.1]{wolff_2006}).

It is pertinent to mention that unsupervised learning in the SP theory (\cite[Chapter 9]{wolff_2006}, \cite[Section 5]{sp_extended_overview}) is quite different from ``Hebbian learning'' as described by Hebb \cite{hebb_1949} and widely adopted in the kinds of artificial neural networks that are popular in computer science.\footnote{See, for example, ``Hebbian theory'', {\em Wikipedia}, \href{http://bit.ly/1sW6ATt}{bit.ly/1sW6ATt}, retrieved 2014-12-19.} By contrast with Hebbian learning, the SP system, like a person, may learn from a single exposure to some situation or event. And, by contrast with Hebbian learning, it takes time to learn a language in the SP system because of the complexity of the search space, not because of any kind of gradual strengthening or ``weighting'' of links between neurons \cite[Section 11.4.4]{wolff_2006}.

\begin{figure}[!htbp]
\centering
\includegraphics[width=0.9\textwidth]{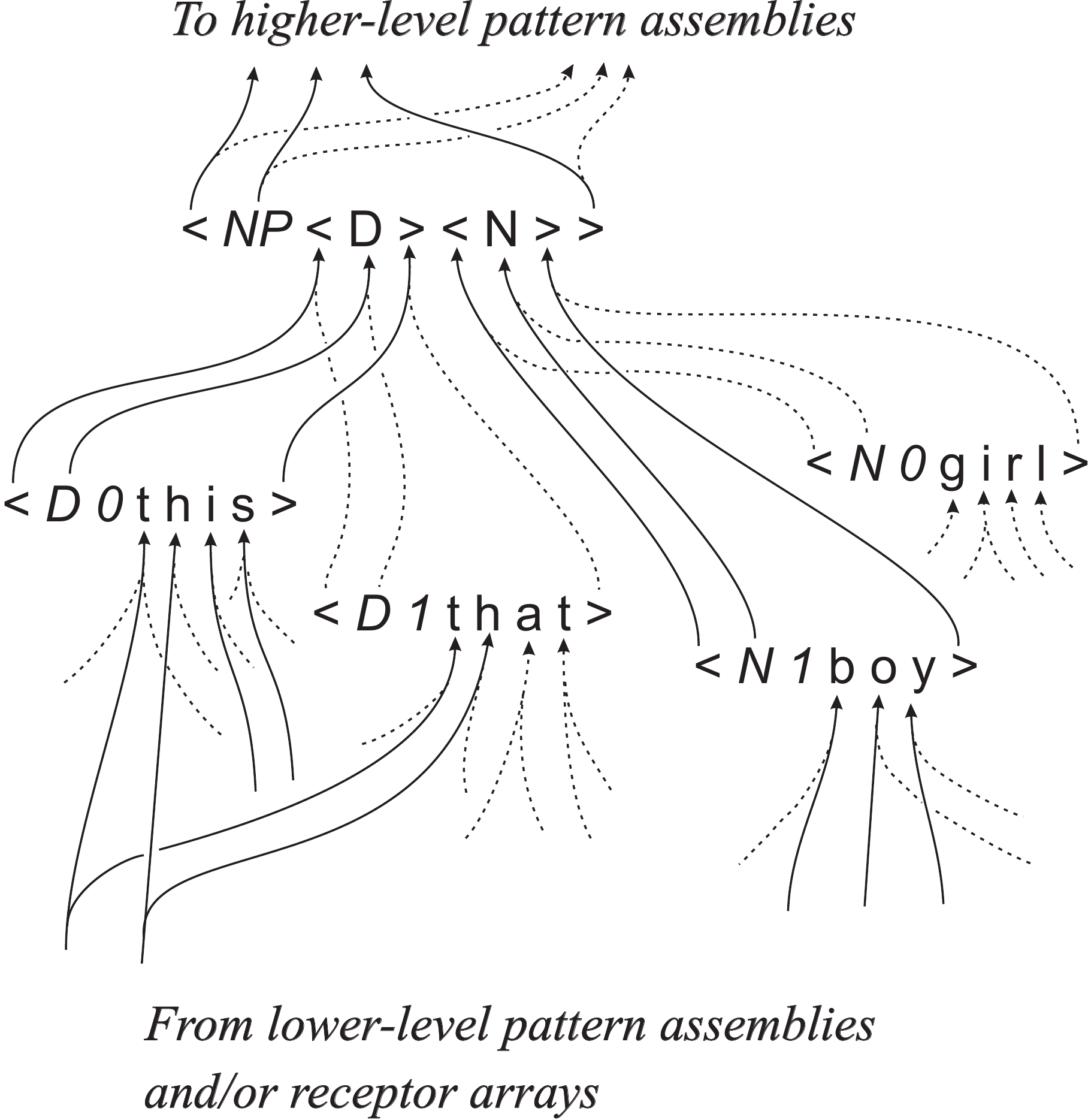}
\caption{Schematic representation of inter-connections amongst pattern assemblies as described in the text. Not shown in the figure are lateral connections within each pattern assembly, and inhibitory connections.}
\label{connections_figure}
\end{figure}

Figure \ref{connections_figure} shows schematically how pattern assemblies may be represented and inter-connected with neurons. Here, each pattern assembly, such as `\texttt{< NP < D > < N > >}', is represented by the sequence of atomic symbols of the corresponding SP pattern. Each atomic symbol, such as `\texttt{<}' or `\texttt{NP}', would be represented in the pattern assembly by one neuron or a small group of inter-connected neurons.\footnote{Not shown in the figure are lateral connections within each pattern assembly and inhibitory connections elsewhere, as outlined in \cite[Sections 11.3.3 and 11.3.4]{wolff_2006}.} Apart from the inter-connections amongst pattern assemblies, the cortex in SP-neural is somewhat like a sheet of paper on which knowledge may be written in the form of neurons.

It is envisaged that any pattern assembly may be `recognised' if it receives more excitatory inputs than rival pattern assemblies, perhaps via a winner-takes-all mechanism \cite[Section 11.3.4]{wolff_2006}. And, once recognised, any pattern assembly may itself be a source of excitatory signals leading to the recognition of higher-level pattern assemblies.

\section{Distinctive features and apparent advantages of the SP system}\label{distinctive_features_and_advantages_appendix}

Information compression and concepts of probability are themes in other research, including research on Bayesian inference, Kolmogorov complexity, deep learning, artificial neural networks, minimum length encoding, unified theories of cognition, natural language processing and more.  The main features that distinguish the SP system from these other areas of research, and apparent advantages compared with these other approaches, are described quite fully in ``The SP theory of intelligence: its distinctive features and advantages'' \cite{sp_alternatives}. The main points are summarised here:

\begin{itemize}

\item {\em Simplification and integration of observations and concepts}. As mentioned above, the SP theory is a unique attempt to simplify and integrate observations and concepts across artificial intelligence, mainstream computing, mathematics, and human perception and cognition. The canvass is much broader than it is, for example, in ``unified theories of cognition''. It has quite a lot to say, for example, about the nature of mathematics \cite{wolff_1993}, \cite[Chapter 10]{wolff_2006}, \cite{sp_foundations}.

\item {\em Simplification and integration of computing systems}. The provision of one simple format for knowledge, and one framework---multiple alignment---for the representation and processing of knowledge, promote an overall simplification of computing systems, including both hardware and software \cite[Section 5]{sp_benefits_apps}. They also promote seamless integration of diverse structures and functions \cite[Section 7]{sp_benefits_apps}.

\item {\em Descriptive and explanatory range}. It appears that the descriptive and explanatory range of the SP system is much wider than any of the alternatives mentioned above. In terms of demonstrable capabilities as well as abstract concepts, it has strengths in areas that include the representation of diverse forms of knowledge (including class hierarchies, part-whole hierarchies, and their seamless integration), unsupervised learning, natural language processing, fuzzy pattern recognition and recognition at multiple levels of abstraction, best-match and semantic forms of information retrieval, several kinds of reasoning (one-step `deductive reasoning', abductive reasoning, probabilistic networks and trees, reasoning with `rules', nonmonotonic reasoning, explaining away, causal reasoning, and reasoning that is not supported by evidence), planning, problem solving, information compression, and aspects of neuroscience and of human perception and cognition.

\item {\em Many potential benefits and applications}. In the spirit of the quote at the start of the Introduction (``There is nothing more practical than a good theory''), the SP theory, with its broad base of support, has many potential benefits and applications, as outlined in Appendix \ref{potential_benefits_applications_appendix}.

\item {\em Strong theoretical underpinnings}. Compared with, for example, artificial neural networks and deep learning, the SP system appears to have much stronger and more coherent theoretical underpinnings. The multiple alignment framework provides a simple but powerful realisation of the concept on which the SP theory is founded: information compression via the matching and unification of patterns.

\item {\em The SP theory is a theory of computing}. Most other research is founded on the idea that computing may be understood in terms of the Universal Turing Machine or equivalent models such as Lamda Calculus or Post's Canonical System. By contrast, {\em the SP theory is itself a theory of computing} \cite[Chapter 4]{wolff_2006}. {\em What is distinctive about the SP theory as a theory of computing is that it provides much of the human-like intelligence that is missing from earlier models}.\footnote{Although Alan Turing saw that computers might become intelligent \cite{turing_1950}, the Universal Turing Machine, in itself, does not tell us how! The SP theory goes some way towards plugging the gap, with potential to do more.}

\item {\em Information compression via the matching and unification of patterns}. In trying to cut through complexities, the SP research programme focuses on a simple, `primitive' idea: that information compression may be understood as a search for patterns that match each other, with the merging or `unification' of patterns that are the same. The potential advantage of this approach, focussing on the simple concept of matching and unifying patterns, is that it can help us avoid old tramlines, and open doors to new ways of thinking.

\item {\em Multiple alignment}. More specifically, information compression via the matching and unification of patterns provides the basis for a concept of {\em multiple alignment}, borrowed and adapted from that concept in bioinformatics. Developing this idea as a framework for the simplification and integration of concepts across a broad canvass has been a major undertaking. {\em Multiple alignment is a distinctive and powerful idea in the SP research programme}.

\item {\em Transparency in the representation and processing of knowledge}. By contrast with sub-symbolic approaches to artificial intelligence (artificial neural networks, deep learning, and related approaches), and notwithstanding objections to symbolic AI,\footnote{See, for example, ``Hubert Dreyfus's views on artificial intelligence'', {\em Wikipedia}, \href{http://bit.ly/1hGHVm8}{bit.ly/1hGHVm8}, retrieved 2014-08-19.} knowledge in the SP system is transparent and open to inspection, and likewise for the processing of knowledge.

\item {\em The DONSVIC principle}. A related point is that unsupervised learning in the SP system is geared to the realisation of the ``DONSVIC'' principle---{\em The Discovery of Natural Structures Via Information Compression} \cite[Section 5.2]{sp_extended_overview}. By contrast with sub-symbolic approaches to artificial intelligence, structures created by learning should, normally, be comprehensible by people.

\item {\em Perception and cognition}. The SP theory draws extensively on research on human and animal perception and cognition, and neuroscience. In particular, an important part of its inspiration is research on the learning of natural language (see \href{http://www.cognitionresearch.org/lang\_learn.html}{www.cognitionresearch.org/lang\_learn.html}).

\item {\em SP-neural}. The SP theory includes proposals---SP-neural---for how abstract concepts in the theory may be realised in terms of neurons and neural processes. The SP-neural proposals (Appendix \ref{sp-neural_appendix}) are significantly different from artificial neural networks as commonly conceived in computer science, and arguably more plausible in terms of neuroscience.

\end{itemize}

\section{Some potential benefits and applications of the SP system}\label{potential_benefits_applications_appendix}

With the SP system, there is a large range of potential benefits and applications, some of which are outlined in the following subsections.

Some potential applications may be developed on relatively short timescales using existing high-performance computers or even ordinary computers. These include the SP system as an intelligent database (Appendix \ref{storage_retrieval_appendix}), and applications in such areas as medical diagnosis (Appendix \ref{medical_diagnosis_appendix}), pattern recognition (Appendix \ref{pattern_recognition_appendix}), information compression (Appendix \ref{ic_app_appendix}), highly-economical transmission of information (Appendix \ref{big_data_appendix}, \cite[Section VIII]{sp_big_data}), bioinformatics (Appendix \ref{bioinformatics_appendix}), and natural language processing (Appendix \ref{syntax_semantics_appendix}).

More radical solutions, that may take longer to develop, include a radically new architecture for computers (Appendix  \ref{new_architecture_appendix}), and developing the full potential of the system in computer vision (Appendix \ref{vision_appendix}), and natural language processing (Appendix \ref{natural_language_processing_appendix}).

\subsection{Simplification and integration of concepts}\label{simplification_integration_concepts_appendix}

\begin{figure}[!htbp]
\centering
\includegraphics[width=0.8\textwidth]{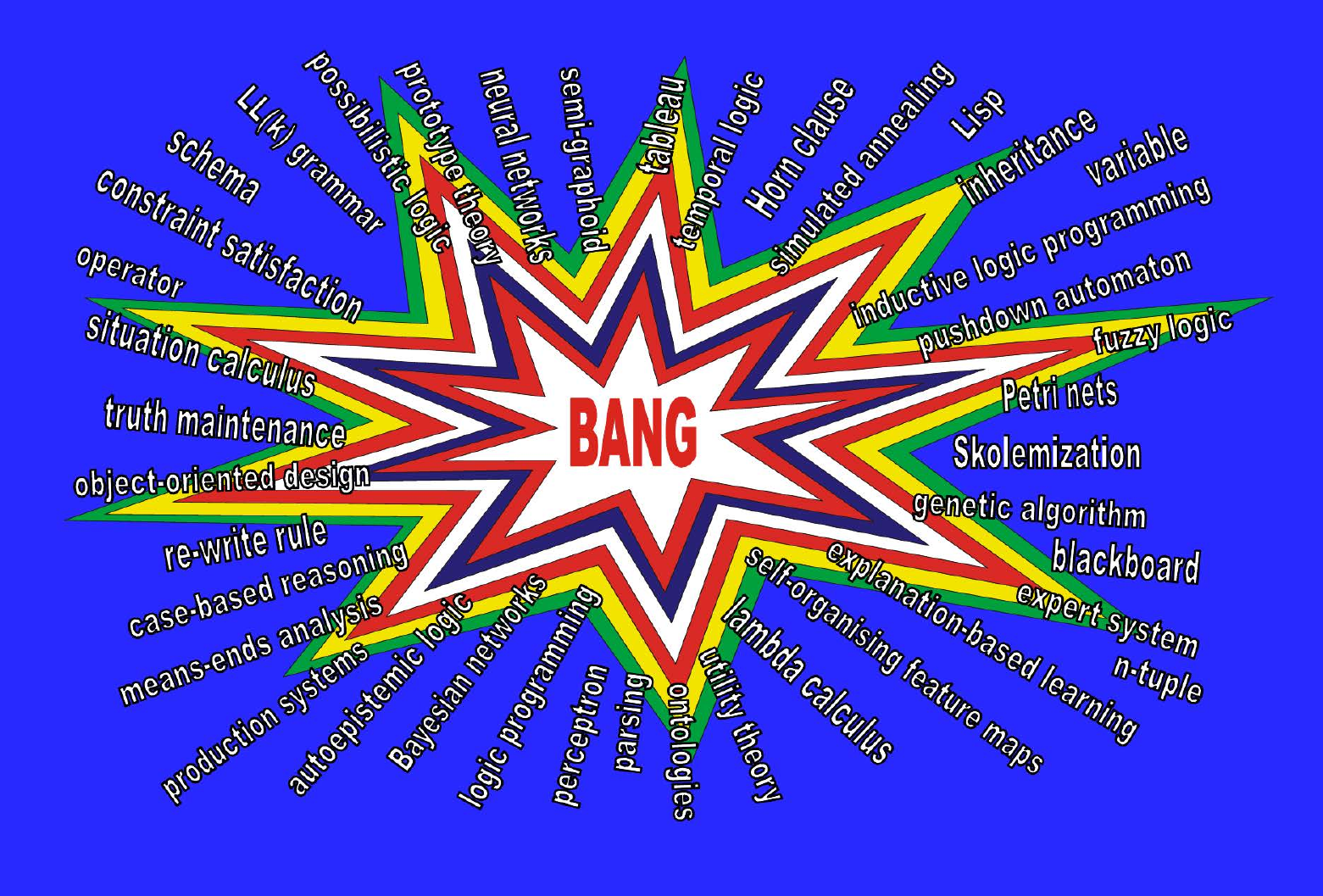}
\caption{Schematic representation of the fragmentation of computer science.}
\label{bang_figure}
\end{figure}

Although some people may argue otherwise, the world of computing suffers from a deep malaise: its fragmentation into a myriad of concepts, shown schematically in Figure \ref{bang_figure}, a myriad of different formalisms and formats for representing knowledge, each with its own mode of processing \cite[Section III]{sp_big_data}, and the extraordinary complexity of many computing systems, especially software.

The quest for simplification and integration in the SP theory accords with Occam's Razor, one of the most widely-accepted principles in science. In terms of that principle, seeking to combine conceptual {\em simplicity} with descriptive and explanatory {\em power}:

\begin{itemize}

\item A relatively simple conceptual framework---multiple alignment---provides an account of a wide range of concepts and phenomena \cite{wolff_2006,sp_extended_overview}.

\item There is potential for very substantial simplification of computing systems, taking hardware and software together \cite[Section 5]{sp_benefits_apps}, and for seamless integration of diverse structures and functions \cite[Sections 2 and 7]{sp_benefits_apps}.

\item Like any theory that simplifies and integrates a good range of observations and concepts, the SP theory promises deeper insights and better solutions to problems than may otherwise be achieved \cite[Section 6]{sp_benefits_apps}, \cite[Section IV-A3]{sp_autonomous_robots}.

\end{itemize}

\subsection{A radically new architecture for computers}\label{new_architecture_appendix}

The SP system has the potential to provide the basis for a radically new architecture for computers, either via the abstract model (with a central role for multiple alignment) or, more likely, via SP-neural. Potential benefits of this new architecture would be human-like versatility and adaptability in computing, substantial simplification of software, and dramatic reductions in the energy demands of computers, and in their size and weight \cite[Section IX]{sp_big_data}.

\subsection{Unsupervised learning}\label{unsupervised_learning_appendix}

The SP theory originates in part from an earlier programme of research on grammatical inference and the unsupervised learning of natural language, with minimum length encoding as a central principle \cite{wolff_1988}, \cite[Section V-A3]{sp_autonomous_robots}. However, meeting the goals of the SP research programme has meant a radical reorganisation of computer models, with the development of multiple alignment as a framework for the simplification and integration of diverse structures and functions \cite[Section V-A4]{sp_autonomous_robots}. The new model demonstrates capabilities in grammatical inference (\cite[Chapter 9]{wolff_2006}, Appendix \ref{usl_grammars_appendix}) and appears to have considerable potential for further development with linguistic, visual, and other kinds of knowledge.

\subsection{Natural language processing}\label{natural_language_processing_appendix}

In addition to the learning of linguistic knowledge (Appendix \ref{unsupervised_learning_appendix}), the SP system has strengths in the parsing of natural language, the production of natural language, and the integration of syntactic and semantic knowledge, as outlined in this section. These aspects of the system are described more fully in \cite[Section 8]{sp_extended_overview} and in \cite[Chapter 5]{wolff_2006}.

\subsubsection{Parsing of natural language}\label{parsing_nl_appendix}

Figure \ref{parsing_figure} shows how, via multiple alignment, a sentence (in row 0) may be parsed in terms of grammatical structures including words (rows 1 to 8).\footnote{Compared with the multiple alignment shown in Figure \ref{class_part_plant_figure}, this multiple alignment is rotated through $90\degree$, replacing columns with rows. The choice between these two styles, which are equivalent, depends largely on what fits best on the page.} It also shows, in row 8, how the system may mark the syntactic dependency between the plural subject of the sentence (`\texttt{Np}') and the plural main verb (`\texttt{Vp}') (see also \cite[Sections 5.4 and 5.5]{wolff_2006}, \cite[Section 8.1]{sp_extended_overview}).

\begin{figure}[!htbp]
\fontsize{07.00pt}{08.40pt}
\centering
{\bf
\begin{BVerbatim}
0                       t h e                a p p l e    s                a r e         s w e e t       0
                        | | |                | | | | |    |                | | |         | | | | |
1                       | | |         N Nr 6 a p p l e #N |                | | |         | | | | |       1
                        | | |         | |              |  |                | | |         | | | | |
2                       | | |    N Np N Nr             #N s #N             | | |         | | | | |       2
                        | | |    | |                        |              | | |         | | | | |
3                  D 17 t h e #D | |                        |              | | |         | | | | |       3
                   |          |  | |                        |              | | |         | | | | |
4            NP 0a D          #D N |                        #N #NP         | | |         | | | | |       4
             |                     |                            |          | | |         | | | | |
5            |                     |                            |  V Vp 11 a r e #V      | | | | |       5
             |                     |                            |  | |           |       | | | | |
6 S Num    ; NP                    |                           #NP V |           #V A    | | | | | #A #S 6
     |     |                       |                                 |              |    | | | | | |
7    |     |                       |                                 |              A 21 s w e e t #A    7
     |     |                       |                                 |
8   Num PL ;                       Np                                Vp                                  8
\end{BVerbatim}
}
\caption{The best multiple alignment created by the SP model with a store of Old patterns like those in rows 1 to 8 (representing grammatical structures, including words) and a New pattern (representing a sentence to be parsed) shown in row 0.}
\label{parsing_figure}
\end{figure}

To create a multiple alignment like the one in the figure, the system needs a grammar of Old patterns, like those shown, one per row, in rows 1 to 8 of the figure. In this example, the patterns represent linguistic structures including words.

Although SP patterns are remarkably simple, it appears that, within the multiple alignment framework, they have at least the expressive power of a context-sensitive grammar \cite[Sections 5.4 and 5.5]{wolff_2006}. As previously noted (Appendix \ref{patterns_and_symbols_appendix}), there is reason to believe that, within the multiple alignment framework, all kinds of knowledge may be represented by SP patterns.

\subsubsection{Production of natural language}

A neat feature of the SP system is that one set of mechanisms may achieve both the analysis or parsing of natural language (Appendix \ref{parsing_nl_appendix}) and the generation or production of sentences. This is explained in \cite[Section 3.8]{wolff_2006} and \cite[Section 4.5]{sp_extended_overview}.

\subsubsection{The integration of syntax and semantics}\label{syntax_semantics_appendix}

The use of one simple format for all kinds of knowledge is likely to facilitate the seamless integration of syntax and semantics. Preliminary examples of how this may be done are shown in \cite[Section 5.7]{wolff_2006}, both for the derivation of meanings from surface forms \cite[Figure 5.18]{wolff_2006} and for the production of surface forms from meanings \cite[Figure 5.19]{wolff_2006}.

It appears that, on relatively short timescales, there is potential via the SP system to create natural language interfaces for such things as internet TVs, DVD recorders and the like, and thus help to overcome much of the difficulty that many people now experience in controlling these things.

\subsubsection{Parallel streams of information}\label{nl_parallel_streams_appendix}

Up to now, most work on natural language within the SP research programme has made the simplifying assumption that language may be represented with a sequence of symbols, as in ordinary text. But with some aspects of natural language such as formants in speech, and the relationship between syntax and semantics, there seem to be parallel streams of information. The way in which such parallelism may be represented and processed with 2D patterns in the SP system is described in \cite[Section IV-B4 and Appendix C]{sp_autonomous_robots}.

\subsection{Pattern recognition}\label{pattern_recognition_appendix}

As described quite fully in \cite[Chapter 6]{wolff_2006} and more briefly in \cite[Section 9]{sp_extended_overview}, the SP system has strengths in several aspects of pattern recognition:

\begin{itemize}

\item It can recognise patterns at multiple levels of abstraction, with the integration of class-inclusion relations and part-whole relations, as shown in the example in Figure \ref{class_part_plant_figure}.

\item It can model ``family resemblance'' or polythetic categories, meaning that recognition does not depend on the presence absence of any particular feature or combination of features.

\item Recognition is robust in the face of errors of omission, commission or substitution in the New pattern or patterns.

\item For any given identification, or any related inference, the SP system may calculate associated probabilities.

\item As a by-product of how recognition is achieved via the building of multiple alignments, the system provides a model for the way in which context may influence recognition.

\end{itemize}

\subsection{Information storage and retrieval, with intelligence}\label{storage_retrieval_appendix}

The SP theory provides a versatile model for database systems, with the ability to accommodate object-oriented structures, as well as relational `tuples', and network and tree models of data \cite{wolff_sp_intelligent_database}. It lends itself most directly to information retrieval in the manner of query-by-example but it appears to have potential to support the use of natural language or query languages such as SQL.

Unlike some ordinary database systems:

\begin{itemize}

\item The storage and retrieval of information is integrated with other aspects of intelligence such as pattern recognition, reasoning, planning, problem solving, and learning---as outlined elsewhere in this document.

\item The SP system provides a simple but effective means of combining class hierarchies with part-whole hierarchies, with inheritance of attributes (Appendix \ref{ma_knowledge_appendix}).

\item It provides for cross-classification with multiple inheritance.

\item There is flexibility and versatility in the representation of knowledge arising from the fact that the system does not distinguish `parts' and `attributes' \cite[Section 4.2.1]{wolff_sp_intelligent_database}.

\item Likewise, the absence of a distinction between `class' and `object' facilitates the representation of knowledge and eliminates the need for a `metaclass' \cite[Section 4.2.2]{wolff_sp_intelligent_database}.

\item SP patterns provide a simpler and more direct means of representing entity-relationship models than do relational tuples \cite[Section 4.2.3]{wolff_sp_intelligent_database}.

\end{itemize}

\subsection{Vision}\label{vision_appendix}

With generalisation of the SP system to accommodate 2D patterns, it has potential to model several aspects of natural vision and to facilitate the development of human-like abilities in artificial vision \cite{sp_vision}. In these connections, the main strengths and potential of the SP system are:

\begin{itemize}

\item Low level perceptual features such as edges or corners may be identified via the multiple alignment framework by the extraction of redundancy in uniform areas in the manner of the run-length encoding technique for information compression.

\item The system may be applied in the recognition of objects and in scene analysis, with the same strengths as in pattern recognition (Appendix \ref{pattern_recognition_appendix}).

\item There is potential for the learning of visual entities and classes of entity and the piecing together of coherent concepts from fragments \cite[Section 5]{sp_vision}.

\item There is potential, via multiple alignment, for the creation of 3D models of objects and of surroundings \cite[Section 6]{sp_vision}.

\item The SP theory provides an account of how we may see things that are not objectively present in an image, how we may recognise something despite variations in the size of its retinal image, and how raster graphics and vector graphics may be unified.

\item And the SP theory has things to say about the phenomena of lightness constancy and colour constancy, ambiguities in visual perception, and the integration of vision with other senses and other aspects of intelligence.

\end{itemize}

\subsection{Reasoning}\label{reasoning_appendix}

As described in quite fully in \cite[Chapters 7 and 10, Section 6.4]{wolff_2006} and more selectively in \cite[Section 10]{sp_extended_overview}, the SP system lends itself to several kinds of reasoning:

\begin{itemize}

\item One-step `deductive' reasoning.

\item Abductive reasoning.

\item Reasoning with probabilistic decision networks and decision trees.

\item Reasoning with `rules'.

\item Nonmonotonic reasoning and reasoning with default values.

\item Reasoning in Bayesian networks, including ``explaining away''.

\item Causal diagnosis.

\item Reasoning which is not supported by evidence.

\item Inheritance of attributes in an object-oriented class hierarchy or heterarchy.

\end{itemize}

There is also potential for spatial reasoning \cite[Section IV-F1]{sp_autonomous_robots} and what-if reasoning \cite[Section IV-F2]{sp_autonomous_robots}.

These several kinds of reasoning may work together seamlessly without awkward incompatibilities, and likewise for how they may integrate seamlessly with such AI functions as unsupervised learning, pattern recognition, and so on \cite[Sections 2, 4, and 7]{sp_benefits_apps}.

For any given inference reached via any of these kinds of reasoning, the SP system may calculate associated probabilities (Appendix \ref{ic_prediction_probabilities_appendix}).

Although the system is fundamentally probabilistic, it may imitate the effect of logic and other `exact' forms of reasoning \cite[Section 10.4.5]{wolff_2006}.

\subsection{Planning and problem solving}

With data about flights between different cities, represented using SP patterns, the SP computer model may find a route between any two cities (if such a route exists) and, if there are alternative routes, it may find them as well \cite[Section 8.2]{wolff_2006}.

Provided they are translated into textual form, the SP computer model can solve geometric analogy problems of the kind found in puzzle books and some IQ tests \cite[Section 8.3]{wolff_2006}, \cite[Section 12]{sp_extended_overview}.

\subsection{Software engineering}\label{software_engineering_appendix}

Although it may not seem obvious at first sight, the multiple alignment framework can model several devices used in ordinary procedural programming, including: {\em procedure}, {\em function}, or {\em subroutine}; {\em variable}, {\em value} and {\em type}; {\em function with parameters}; {\em conditional statement}; and the means of repeating operations such as {\em repeat ...~until} or {\em do ...~while} \cite[Section 6.6.1]{sp_benefits_apps}. In accordance with good practice in software engineering, the SP system facilitates the integration of `programs' with `data'. And as previously noted (Appendix \ref{ma_knowledge_appendix}), the SP system supports object-oriented concepts such as class hierarchies with inheritance of attributes.

In \cite[Section 6.6.3]{sp_benefits_apps}, it is suggested that, since SP patterns at the `top' level are independent of each other, they may serve to model processes that may run in parallel. Now it appears that a better option is to model parallel processes as parallel streams of information, represented in 2D SP patterns as described in \cite[Appendix C]{sp_autonomous_robots}. The advantage of this latter scheme is that it provides the means of showing when two or more events occur at the same time and, more generally, the relative timings of events.

Within the SP system, these structures and mechanisms may serve in the representation and processing of sequential and parallel procedures from the real world such as those required for cooking a meal, organising a party, going shopping, and so on.

Potential benefits in software engineering include the elimination of compiling or interpretation, automatic programming, benefits in verification and validation, and helping to overcome the problem of technical debt \cite[Section 6.6]{sp_benefits_apps}.

\subsection{Mathematics}

Aspects of mathematics may be understood in terms of some basic techniques for information compression: {\em chunking-with-codes}, {\em schema-plus-correction}, and {\em run-length coding} \cite{sp_foundations}, and some features of mathematics may be modelled in the SP system \cite[Chapter 10]{wolff_2006}.

On the strength of this evidence and some other considerations, there is reason to believe that all of mathematics may be understood in terms of information compression. There appear to be considerable implications for mathematics, and also for science---because of the importance of mathematics as a language of science.

\subsection{Human perception and cognition, and neuroscience}

Part of the inspiration for the SP theory has been earlier research on the role of information compression in the workings of brains and nervous systems \cite{attneave_1954,barlow_1959,barlow_1969}, and a programme of research on the learning of a first language or languages \cite{wolff_1988}. Thus there is reason to believe that the SP theory may help to illuminate human perception and cognition.

Since the converse is true---that insights into the nature of human perception and cognition may help to inform the development of the SP theory---the two things are often considered together, as can be seen in \cite{wolff_2006,sp_vision} and other writings about the SP system.

As outlined in Appendix \ref{sp-neural_appendix}, the SP theory includes the proposal---called {\em SP-neural}---that abstract concepts in the theory may be realised in terms of neurons and their interconnections \cite[Chapter 11]{wolff_2006}.

\subsection{Other potential benefits and applications}

The versatility of the SP system may be seen not only in the areas outlined above but in potential benefits and applications summarised in the following subsections.

\subsubsection{Big data}\label{big_data_appendix}

The SP system may help to solve nine problems associated with big data \cite{sp_big_data}. In brief, these problems and their potential solutions are:

\begin{itemize}

\item {\em Overcoming the problem of variety in big data}. Harmonising diverse kinds of knowledge, diverse formats for knowledge, and their diverse modes of processing, via a universal framework for the representation and processing of knowledge (UFK).

\item {\em Learning and discovery}. The unsupervised learning or discovery of `natural' structures in data.

\item {\em Interpretation of data}. The system has strengths in areas such as pattern recognition, information retrieval, parsing and production of natural language, translation from one representation to another, several kinds of reasoning, planning and problem solving.

\item {\em Velocity: analysis of streaming data}. The SP system lends itself to an incremental style, assimilating information as it is received, much as people do.

\item {\em Volume: making big data smaller}. Reducing the size of big data via lossless compression can yield several benefits.

\item {\em Transmission of data}. There is potential for substantial economies in the transmission of data by judicious separation of `encoding' and `grammar'.

\item {\em Energy, speed, and bulk}. There is potential for big cuts in the use of energy in computing, for greater speed of processing with a given computational resource, and for corresponding reductions in the size and weight of computers.

\item {\em Veracity: managing errors and uncertainties in data}. The SP system can identify possible errors or uncertainties in data, suggest possible corrections or interpolations, and calculate associated probabilities.

\item {\em Visualisation}. Knowledge structures created by the system, and inferential processes in the system, are all transparent and open to inspection. They lend themselves to display with static and moving images.

\end{itemize}

Considering these proposed solutions collectively, and in several cases individually, it appears that there are no alternatives that can rival the potential of what is described in \cite{sp_big_data}.

\subsubsection{Autonomous robots}\label{autonomous_robots_appendix}

The SP system may help in the design of autonomous robots, meaning robots that do not depend on external intelligence or power supplies, are mobile, and are designed to exhibit as much human-like intelligence as possible \cite{sp_autonomous_robots}. The three main areas where it may make a contribution---problems with the efficiency and bulkiness of computers; developing human-like versatility in robots; and developing human-like adaptability in robots---are summarised in Section \ref{robotics_autonomous_section}.

\subsubsection{Medical diagnosis}\label{medical_diagnosis_appendix}

The way in which the SP system may be applied in medical diagnosis is described in \cite{wolff_medical_diagnosis}. The expected benefits of the SP system in that area of application include:

\begin{itemize}

\item A format for representing diseases that is simple and intuitive.

\item An ability to cope with errors and uncertainties in diagnostic information.

\item The simplicity of storing statistical information as frequencies of occurrence of diseases.

\item The system provides a method for evaluating alternative diagnostic hypotheses that yields true probabilities.

\item It is a framework that should facilitate the unsupervised learning of medical knowledge and the integration of medical diagnosis with other AI applications.

\end{itemize}

The main emphasis in \cite{wolff_medical_diagnosis} is on medical diagnosis as pattern recognition. But the SP system may
also be applied to causal diagnosis \cite[Section 7.9]{wolff_2006}, \cite[Section 10.5]{sp_extended_overview} so that it may be possible, for example, to reason that ``The patient's fatigue may be caused by anemia which may be caused by a shortage of iron in the diet''.

\subsubsection{Bioinformatics}\label{bioinformatics_appendix}

Because of the central importance of multiple alignment in the SP system, and because of the importance of that concept in bioinformatics, there is clear potential for the SP system to find applications in that area \cite[Section 6.10.2]{sp_benefits_apps}. Multiple alignment as it has been developed in the SP system has potential advantages compared with multiple alignment as it has been developed for bioinformatics.

Another potential advantage of the SP system is its capabilities and potential in unsupervised learning (Appendices \ref{usl_grammars_appendix} and \ref{unsupervised_learning_appendix}, and sources referenced there):

\begin{itemize}

\item They have potential to discover recurrent patterns of arbitrary size within DNA sequences, amino acid sequences, and the like---including sequences that are discontinuous, jumping over intervening structures.

\item They have potential to discover disjunctive classes of entity, with the context in which they are embedded.

\item With the right kind of data, there is potential to discover structures and associations within and between DNA, proteins, enzymes, and associated biochemical processes.

\item Likewise, there is potential to discover associations between diseases and their symptoms on the one hand, and biochemical structures and processes on the other.

\end{itemize}

\subsubsection{Information compression}\label{ic_app_appendix}

Since information compression is central in the workings of the SP system (Appendix \ref{information_compression_appendix}), there is reason to believe that an industrial-strength version of the system will be useful in that area of application.

\bibliographystyle{plain}
% \bibliography{latex_references}

\end{document}